# Discriminative Local Sparse Representation by Robust Adaptive Dictionary Pair Learning

Yulin Sun, Zhao Zhang, *Senior Member, IEEE*, Weiming Jiang, Zheng Zhang, *Member, IEEE*, Li Zhang, *Member, IEEE*, Shuicheng Yan, *Fellow, IEEE*, and Meng Wang, *Senior Member, IEEE*

*Abstract*— In this paper, we propose a structured Robust Adaptive Dictionary Pair Learning (RA-DPL) framework for the discriminative sparse representation learning. To achieve powerful representation ability of the available samples, the setting of RA-DPL seamlessly integrates the robust projective dictionary pair learning, locality-adaptive sparse representations and discriminative coding coefficients learning into a unified learning framework. Specifically, RA-DPL improves existing projective dictionary pair learning in four perspectives. First, it applies a sparse $l_{2,1}$-norm based metric to encode the reconstruction error to deliver the robust projective dictionary pairs, and the $l_{2,1}$-norm has the potential to minimize the error. Second, it imposes the robust $l_{2,1}$-norm clearly on the analysis dictionary to ensure the sparse property of the coding coefficients rather than using the costly $l_0/l_1$-norm. As such, the robustness of the data representation and the efficiency of the learning process are jointly considered to guarantee the efficacy of our RA-DPL. Third, RA-DPL conceives a structured reconstruction weight learning paradigm to preserve the local structures of the coding coefficients within each class clearly in an adaptive manner, which encourages to produce the locality preserving representations. Fourth, it also considers improving the discriminating ability of coding coefficients and dictionary by incorporating a discriminating function, which can ensure high intra-class compactness and inter-class separation in the code space. Extensive experiments show that our RA-DPL can obtain superior performance over other state-of-the-arts.

## I. Introduction

EFFECTIVE image representation and classification via dictionary learning (DL) have received much attention in recent years and have also been successfully applied to a variety of real-world emerging applications in a wide range of areas, such as computer vision [13], image denoising and compression [14], visual image classification [1-12][38-41], etc. Technically, DL aims at computing the sparse representations (SR) of samples by a linear combination of dictionary atoms, so the properties and superiority of learned dictionary will play a crucial role for SR [15]. Wright *et al.* [4] have used the entire training set as a dictionary to represent the samples yielding an impressive face recognition result, but note that two drawbacks of using such a dictionary may potentially decrease its performance. First, real application data are usually corrupted by various noise and errors [2]; Second, this kind of dictionary usually has a large size to make the process of coefficients coding inefficient [1-2]. To address these issues, lots of efforts have devoted to the research of learning the compact and over-complete dictionaries in the area of representation learning [1-12][47-54].

The existing compact dictionary learning frameworks can be roughly categorized into unsupervised and discriminant groups. The unsupervised methods do not apply any prior label information of training data and aims at minimizing a reconstruction residual over input data to produce dictionaries [3-5], among which *K-Singular Value Decomposition* (K-SVD) [3] is one of the most representative unsupervised DL methods. However, it is incapable of handling the classification task since it only expects the learned dictionary to be able to represent data effectively [7]. In contrast, by taking label information of data into consideration, many discriminative algorithms have been recently proposed for enhancing the representation and classification results. For the discriminative DL, one popular strategy is to obtain an overall dictionary for all classes while forcing the resulting coding coefficients to be discriminative, such as *Discriminative K-SVD* (D-KSVD) [6] and *Label Consistent K-SVD* (LC-KSVD) [1] are two classical algorithms. D-KSVD incorporates the classification error term into K-SVD model to enhance the classification result, while LC-KSVD further incorporates the label consistency to D-KSVD for ensuring the discrimination of learned coding coefficients. The other one popular strategy is the structured discriminative DL that aims to obtain category-specific dictionaries and encourage each sub-dictionary to correspond to a single class, such as *Fisher Discrimination Dictionary Learning* (FDDL) [7], *Projective Dictionary Pair Learning* (DPL) [9], *Dictionary Learning with Structured Incoherence* (DLSI) [8], *Analysis Discriminative Dictionary Learning* (ADDL) [11], *Low-rank Shared Dictionary Learning* (LRSDL) [24], etc. FDDL enforces a Fisher criterion on the representation coefficients and residual to obtain a structured dictionary and enables the coefficients to deliver small intra-class scatter and large inter-class scatter. DPL obtains an extra analysis dictionary over traditional synthesis dictionary learning for representation learning and classification. DLSI incorporates an incoherence promoting term to ensure the independence among sub-dictionaries. ADDL further extends DPL to jointly learn the structured uncorrelated dictionaries and a linear analysis classifier, and uses $l_{2,1}$-norm regularization to deliver sparse coefficients due to efficiency. In this paper, the structured DL mechanism will be further investigated.

It is worth noting that existing structured DL algorithms still suffer from some shortcomings that may lead to inferior

---

Y. Sun, W. Jiang and L. Zhang are with the School of Computer Science and Technology, Soochow University, Suzhou 215006, China. (e-mails: daitusun@gmail.com, csjiangwm@outlook.com, zhangliml@suda.edu.cn)

Zhao Zhang is with School of Computer Science & School of Artificial Intelligence, Hefei University of Technology, Hefei, China; also with the School of Computer Science and Technology, Soochow University, Suzhou 215006, China. (e-mail: cszzhang@gmail.com)

M. Wang is with School of Computer Science & School of Artificial Intelligence, Hefei University of Technology, Hefei, China. (e-mail: eric.mengwang@gmail.com)

Z. Zhang is with Bio-Computing Research Center, Harbin Institute of Technology, Shenzhen, China, and is also with Pengcheng Labtorary, Shenzhen, China. (e-mail: darrenzz219@gmail.com)

S. Yan is with YITU Technology; also with the National University of Singapore, Singapore. (e-mail: shuicheng.yan@yitu-inc.com)

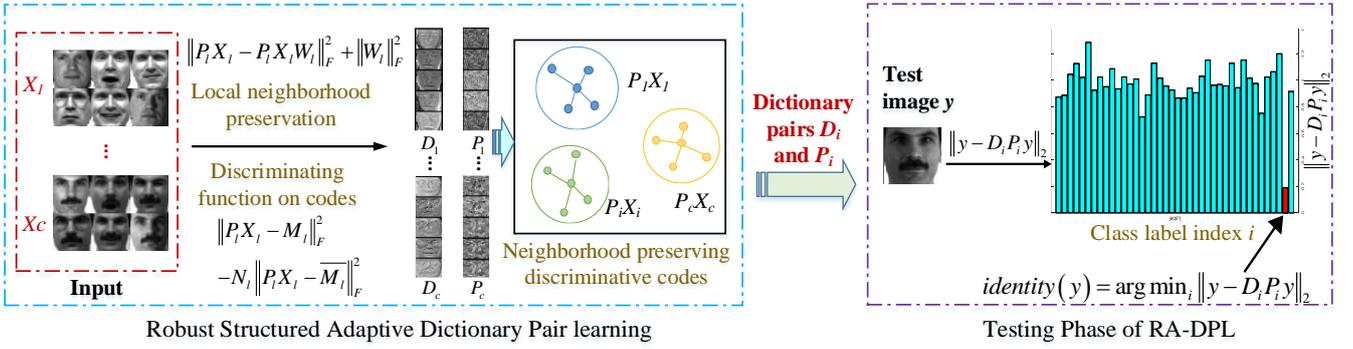

**Fig.1**: Flow-diagram of our proposed RA-DPL framework for image recognition.

performance. First, FDDL, DPL, DLSI, ADDL and LRSDL mentioned-above aim to encode the reconstruction error and perform SR using the Frobenius-norm that is very sensitive to noise and outliers in given data [17][43-44]. Since most real data inevitably contain noise, errors or even corruptions, it would be better to enable the reconstruction metric to be robust. Second, these methods cannot keep the neighbourhood information of the intra-class coding coefficients, especially in an adaptive manner. As a result, they are unable to obtain the locality-preserving coefficients by DL. Third, the intra-class compactness and inter-class separation over the coding coefficients are not well explored in most existing methods. Although FDDL uses the Fisher criterion to make the coefficients of each class $l$ close to the mean of the coefficients in class $l$, it still cannot explicitly ensure the coefficients of class $l$ to be far away from the mean of other classes at the same time. Because the Fisher criterion only encourages the mean of the coefficients in class $l$ to be far away from the total mean of the coefficients of all classes.

In this paper, we therefore investigate the robust adaptive category-specific dictionary learning problem and propose new model to enhance the representation and classification abilities. The major contributions are shown as follows:

(1) Technically, a structured Robust Adaptive Projective Dictionary Pair Learning (RA-DPL) framework is proposed for learning discriminative sparse representation. RA-DPL is based on existing DPL, but it overcomes the drawbacks of DPL and inherits its merits at the same time. Specifically, RA-DPL improves the representation ability by jointly enhancing the robustness of sparse reconstruction and analysis sub-dictionary learning to noise and errors in given data by the robust analysis dictionary learning, preserving the local neighbourhood by adaptive weight learning, and obtaining discriminant sparse coefficients. By the robust structured DL, RA-DPL can compute a synthesis sub-dictionary $D_i$ of each class $i$ separately to reconstruct the data of the same class, and meanwhile learn a robust analysis sub-dictionary $P_i$ each class $i$ separately to extract the discriminative sparse representations from the samples of the same class as well. The relationship analysis between our formulation and other related methods shows that several existing methods can be regarded as the special cases of our formulation.

(2) To enhance the robustness of the sparse reconstruction and analysis sub-dictionary learning, RA-DPL employs the sparse $l_{2,1}$-norm to encode the sparse reconstruction error [17][43-44]. Moreover, as the $l_{2,1}$-norm based metric can make more rows of the reconstruction error matrix to be zeros theoretically [43-44], it can enable the reconstruction error to be as small as possible to deliver reliable and robust projective dictionary pairs and representations. In addition, RA-DPL also leverages the $l_{2,1}$-norm on the analysis dictionary to extract group sparse coefficients from data, due to its efficient optimization in training phase, compared with the widely-used $l_0$-/ $l_1$-norm based formulation.

(3) To capture the locality manifold structures of coding coefficients and make the representation more accurately, RA-DPL incorporates the adaptive reconstruction weighting into the robust analysis dictionary learning to preserve the local neighbourhood of coefficients within each class, such that the discriminating ability of the associated dictionary is also potentially improved. The adaptive weighting strategy is mainly driven by minimizing the adaptive reconstruction error, where the reconstruction weights are clearly shared in the data space and sparse coding space.

(4) Our RA-DPL also considers the discriminative sparse representation. Specifically, we design a discriminative coefficient learning function to highlight the intra-class compactness and inter-class separation over the coding coefficients. The discriminative function can clearly make the coding coefficients of each class $l$ be close to the mean of the same class, and meanwhile can enable the coding coefficients of the class $l$ to be far away from the mean of other classes at the same time, which makes our RA-DPL clearly different from yet superior to existing methods.

The paper is outlined as follows. Section II reviews the related work briefly. Sections III presents RA-DPL. Section IV shows the simulation results. The relationship analysis between our RA-DPL and other related algorithms is shown in Appendix I. Finally, Section V concludes the paper.

## II. RELATED WORK

In this section, we briefly review the related methods that are closely related to our formulation.

### A. Overall Dictionary Learning (ODL)

Let $X = [x_1, \cdots x_l, \cdots x_N] \in \mathbb{R}^{n \times N}$ be a set of training samples from $c$ classes, where $n$ is the original dimensionality and $N$ is the number of samples. Then, ODL learns a reconstructive dictionary $D$ of $K$ atoms to deliver the sparse representation $S$ over the data $X$ by the following general problem:

$$\langle D, S \rangle = \arg\min_{D,S} \| X - DS \|_F^2 + \lambda \| S \|_p, \quad (1)$$

where $\| \bullet \|_F^2$ denotes the Frobenius-norm, $\| X - DS \|_F^2$ is the reconstruction error over $X$ for SR, $D = [d_1, \cdots d_K] \in \mathbb{R}^{n \times K}$ is a dictionary, $S = [s_1, \cdots s_N] \in \mathbb{R}^{K \times N}$ is the coding coefficient matrix over $X$ and $\lambda > 0$ is a scalar constant. $\| S \|_p$ is $l_p$-norm regularization, where $p = 0$ or $1$ is widely-used to ensure the sparse property of *S*, i.e., $l_0$-norm or $l_1$-norm, but such an operation usually incurs a heavy computation burden. To extend ODL for classification, two effective strategies are widely-used, i.e., *sparse representation based classification* (SRC) by residual minimization [4] and the label fitting by embedding. To classify each new sample $x_{new}$ by residual minimization, its coefficient vector $s_{new}$ is firstly computed using well-trained dictionary *D*. Then, the new sample $x_{new}$ can be classified by minimizing the following residual:

$$identity(x_{new}) = \arg\min_l \|x_{new} - D\delta_l(s_{new})\|_2, \quad (2)$$

where $\delta_l(s_{new})$ is a vector whose nonzero entries in $s_{new}$ are associated with class $l \in \{1, \cdots c\}$. That is, the label of $x_{new}$ is assigned to the class with minimum residual [4]. In contrast, the label fitting based scheme obtains a dictionary *D* and a linear multi-class classifier $W \in \mathbb{R}^{c \times K}$ over the coding coefficients jointly, e.g., [7-9][11][24]. A unified problem for learning *D* and *W* jointly can be formulated as

$$\langle D, S, W \rangle = \arg\min_{D,S,W} \|X - DS\|_F^2 + \lambda \|S\|_p + \sum_i \Psi\{h_i, f(s_i, W)\}, \quad (3)$$

where $\Psi$ indicates the classification loss function and $h_i$ is the pre-defined label of each $x_i$. Thus, $x_{new}$ can be classified by embedding its coefficient vector $s_{new}$ into trained classifier *W*. It is worth noting that both methods classify each new sample based on the sparse coding coefficients, but an extra time-consuming sparse reconstruction process is usually needed for each new test sample for classification.

### B. Structured Dictionary Learning (SDL)

In the supervised cases, the training data matrix *X* usually has samples from *c* classes, i.e., $X = [X_1, \cdots X_l, \cdots X_c] \in \mathbb{R}^{n \times N}$, where $X_l \in \mathbb{R}^{n \times N_l}$ is a sub-matrix according to the class *l*, and $N_l$ is the number of samples in the class *l*, i.e., $\sum_{l=1}^c N_l = N$. Then, the structured DL can be performed based on each $X_l$. Subsequently, we briefly review two related structured DL algorithms, i.e., DPL [9] and FDDL [7].

**DPL.** To avoid the heavy burden caused by using costly $l_0$ or $l_1$-norm, DPL proposes to learn a synthesis dictionary *D* and an analysis dictionary *P* for group SR by solving the following problem to avoid the costly constraint:

$$\langle P, D \rangle = \arg\min_{P,D} \sum_{l=1}^c \|X_l - D_l P_l X_l\|_F^2 + \lambda \|P_l \overline{X_l}\|_F^2, \quad s.t. \|d_i\|_2^2 \leq 1, \quad (4)$$

where $\overline{X_l}$ is the complementary data matrix of $X_l$ in *X*, i.e., excluding $X_l$ itself from *X*. $D_l = [d_1, \cdots d_{k_l}] \in \mathbb{R}^{n \times k_l}$ with $k_l$ atoms denotes the synthesis sub-dictionary for each subject class *l*, $P_l \in \mathbb{R}^{k_l \times n}$ is an analysis sub-dictionary for each class *l*, $D = [D_1, \cdots D_l, \cdots D_c] \in \mathbb{R}^{n \times K}$ and $P = [P_1; \cdots P_l; \cdots P_c] \in \mathbb{R}^{K \times n}$. The constraint $\|d_i\|_2^2 \leq 1$ prevents the large values of *D* and avoids the trivial solution $P_l = 0$ to make the computation stable. Note that DPL regularizes the group sparsity on the coding coefficients *PX* (i.e. *PX* is nearly block-diagonal).

**FDDL.** FDDL aims to obtain a structured dictionary and forces the coding coefficients to deliver small within-class scatter and large between-class scatter by minimizing the following Fisher criterion based cost function:

$$J(D, S) = \frac{1}{2} \sum_{l=1}^c r(X_l, D, S_l) + \lambda_1 \|S\|_1 + \frac{\lambda_2}{2} g(S), \quad (5)$$

where $J(D, S)$ is the discriminative data fidelity term, $g(S)$ is the Fisher-criterion based discriminative coefficients term, and $\|S\|_1$ is the $l_1$-norm to ensure the sparsity of coefficients. Note that the terms $r(X_l, D, S_l)$ and $g(S)$ are defined as

$$r(X_l, D, S_l) = \|X_l - DS_l\|_F^2 + \|X_l - D_l S_l^l\|_F^2 + \sum_{j \neq l} \|D_j S_l^j\|_F^2$$

$$g(S) = \sum_{l=1}^c \left( \sum_{i=1}^{N_l} \|s_l^i - m_l\|_2^2 - N_l \|m_l - m\|_F^2 \right) + \|S\|_F^2$$

where $m_l$ and $m$ are the mean vectors over $S_l$ and $S$, respectively. Based on the regularization $\|S\|_F^2$, the above model can be ensured to be convex with respect to *S*.

## III. DISCRIMINATIVE LOCAL SPARSE REPRESENTATIONS BY ROBUST ADAPTIVE DICTIONARY PAIR LEARNING

### A. Objective Function

We describe the objective function of RA-DPL. To improve the representation and discriminating abilities, our RA-DPL performs the robust structured dictionary pair learning by minimizing the sparse $l_{2,1}$-norm based reconstruction error $\sum_{l=1}^c \|X_l^T - X_l^T P_l^T D_l^T\|_{2,1} + \alpha \|P_l^T\|_{2,1}$, where $\|P_l^T\|_{2,1}$ can potentially produce the sparse coding coefficients $P_l X_l$ and make the embedding robust to noise and outliers. By regularizing the $l_{2,1}$-norm on the reconstruction error, one can also implicitly minimize the reconstruction error as much as possible, since the $l_{2,1}$-norm can enforce the error matrix to be sparse in rows [43-44]. To enable the analysis sub-dictionary $P_l$ to project the training data of class $j$ ($j \neq l$) to a nearly null space, i.e., $P_l X_j \approx 0, \forall j \neq l$, a constraint $\sum_{l=1}^c \|P_l \overline{X_l}\|_F^2$ is also applied similarly as [9][11]. For the locality-adaptive SR, RA-DPL adds an adaptive structured reconstruction weight learning function to encode the neighbourhood relationship within one subject class by minimizing the neighbourhood reconstruction error $\sum_{l=1}^c \left( \|X_l - X_l W_l\|_F^2 + \|P_l X_l - P_l X_l W_l\|_F^2 + \|W_l\|_F^2 \right)$, where $W_l$ is the reconstruction weight matrix over class *l*. For the discriminative codes learning, RA-DPL introduces a discriminating function $\sum_{l=1}^c \|P_l X_l - M_l\|_F^2 - N_l \|P_l X_l - \overline{M_l}\|_F^2$ to enhance the compactness of intra-class coefficients and separation of inter-class coefficients so that discriminative representations can be obtained. Therefore, the objective function of our RA-DPL can be defined as

$$\langle D, P, W \rangle = \arg\min_{D,P,W} \sum_{l=1}^c \|X_l^T - X_l^T P_l^T D_l^T\|_{2,1} + \alpha \left( \|P_l \overline{X_l}\|_F^2 + \|P_l^T\|_{2,1} \right)$$
$$+ \beta \left( \|X_l - X_l W_l\|_F^2 + \|P_l X_l - P_l X_l W_l\|_F^2 + \|W_l\|_F^2 \right) \quad , (6)$$
$$+ \lambda \left( \|P_l X_l - M_l\|_F^2 - N_l \|P_l X_l - \overline{M_l}\|_F^2 \right)$$

$$s.t. \ e^T D_l = e^T, P_l X_l \geq 0, \ diag(W_l) = 0$$

where $M_l = [m_{l,1}, m_{l,2}, ..., m_{l,N_l}] \in \mathbb{R}^{k \times N_l}$ is the mean matrix over the coefficients $P_l X_l$ of class *l*, $m_{l,i}$ is the mean vector based on $P_l X_l$, $\overline{M_l} = [\psi_{l,1}, \psi_{l,2}, ..., \psi_{l,N_l}] \in \mathbb{R}^{k \times N_l}$ is the mean matrix based on $P\overline{X_l}$, $\psi_{l,i}$ is the mean vector over $P\overline{X_l}$, and the sum-to-one constraint $e^T D_l = e^T$ can similarly normalize the atoms to avoid the trivial solution $P_l = 0$ as the constraint $\|d_i\|_2^2 \leq 1$. Note that minimizing $\|P_l X_l - M_l\|_F^2$ can ensure the coding coefficients $P_l X_l$ of class *l* to be close to its own mean as much as possible, while maximizing the term $N_l \|P_l X_l - \overline{M_l}\|_F^2$ can clearly make the coefficients of class *l* to be far away from the coefficients of other subject classes. $diag(W_l) = 0$ is added to avoid the trivial solution $W_l = I$ and $P_l X_l \geq 0$ can ensure the nonnegative properties of the embedded coding coefficients $P_l X_l$. $\alpha$, $\beta$ and $\lambda$ are positive parameters to balance the importance of different terms. For easy understanding of our method, the flow-diagram of our RA-DPL framework is illustrated in Fig.1, which illustrates the training process by robust adaptive projective dictionary pair learning and the test phase.

The unified framework of our RA-DPL can be simplified into the following two separable steps:

*1) Robust structured adaptive dictionary pair learning*

Given the structured adaptive reconstruction weight matrix $W$, we can use the following reduced sub-problem from Eq. (6) for the robust discriminative dictionary learning:

$$\langle D,P\rangle = \arg\min_{D,P}\sum_{l=1}^{c}\left\|X_l^T - X_l^T P_l^T D_l^T\right\|_{2,1} + \alpha\left(\left\|P_l\overline{X_l}\right\|_F^2 + \left\|P_l^T\right\|_{2,1}\right)$$
$$+ \beta\left\|P_lX_l - P_lX_lW_l\right\|_F^2 + \lambda\left(\left\|P_lX_l - M_l\right\|_F^2 - N_l\left\|P_lX_l - \overline{M_l}\right\|_F^2\right), \quad (7)$$
$$s.t.\ e^T D_l = e^T, P_l X_l \geq 0$$

from which we can achieve a synthesis dictionary $D$ and an analysis dictionary $P$. It should be noted that FDDL also involves a Fisher-criterion based discriminative coefficients learning term $g(S)=\sum_{l=1}^{c}\left(\sum_{i=1}^{N_l}\left\|s_l^i - m_l\right\|_2^2 - N_l\left\|m_l - m\right\|_F^2\right)+\left\|S\right\|_F^2$ for jointly achieving the inter-class discrimination and intra-class compactness. But note that the discriminating function $\left\|P_lX_l - M_l\right\|_F^2 - N_l\left\|P_lX_l - \overline{M_l}\right\|_F^2$ in RA-DPL is different from the Fisher-criterion based discriminative term $g(S)$ of FDDL. Since FDDL aims at maximizing the difference $\left\|m_l - m\right\|_F^2$ between each class mean $m_l$ and the total mean $m$, but it cannot potentially ensure that the mean $m_l$ of class $l$ to be far away from the mean $m_j (l \neq j)$ of class $j$. In contrast, RA-DPL can clearly ensure the mean matrix $M_l$ of class $l$ to be far away from the mean of other classes by maximizing $N_l\left\|P_lX_l - \overline{M_l}\right\|_F^2$. After updating the analysis dictionary $P$ at each time, one can use it for learning the weights.

*2) Adaptive structured reconstruction weight learning*

When the analysis dictionary $P$ is known, we can compute the adaptive reconstruction weights by preserving the local neighborhood relationship of the training data jointly in the coefficients coding space. In this way, we have the following reduced problem for adaptive reconstruction weighting:

$$W = \arg\min_W \sum_{l=1}^{c}\beta\left(\left\|X_l - X_lW_l\right\|_F^2 + \left\|P_lX_l - P_lX_lW_l\right\|_F^2 + \left\|W_l\right\|_F^2\right), \quad (8)$$
$$s.t.\ diag(W_l) = 0$$

where $\left\|X_l - X_lW_l\right\|_F^2$ denotes the reconstruction error over $X_l$ and $\left\|P_lX_l - P_lX_lW_l\right\|_F^2$ denotes the reconstruction error based on $P_lX_l$. Clearly, the neighborhood information of training data of each class can be kept in the sparse coding space so that both discriminating and locality preserving properties can be concurrently encoded. After the adaptive reconstruction weights are obtained, we can return it for robust adaptive dictionary pair learning by Eq.(7). Note that an early version of this work has been presented in [45]. This paper further integrates the discriminating function on the coding coefficients to deliver discriminative sparse representations, details the formulation analysis, provides the convergence analysis, time complexity analysis and relationship analysis. Moreover, we conduct a thorough experimental evaluation on the tasks of image representation and recognition.

*B. Optimization*

In this section, we present the optimization procedures of our RA-DPL. Because our model involves several variables and the optimization of variables depends on each other, it is still challenging to give all variables an optimal solution jointly. To this end, we propose to solve the problem by an alternative learning, i.e., updating one variable at each time by fixing others. As the optimization problem of RA-DPL in Eq.(6) is generally non-convex, we add a variable matrix $S$ ($S_l^T \approx X_l^T P_l^T$) to relax the problem as

$$\langle D,P,W,S\rangle = \min_{D,P,W,S}\sum_{l=1}^{c}\left(\left\|X_l^T - S_l^T D_l^T\right\|_{2,1}\right) + \left\|S_l^T - X_l^T P_l^T\right\|_{2,1}$$
$$+ \alpha\left(\left\|P_l\overline{X_l}\right\|_F^2 + \left\|P_l^T\right\|_{2,1}\right)$$
$$+ \beta\left(\left\|X_l - X_lW_l\right\|_F^2 + \left\|P_lX_l - P_lX_lW_l\right\|_F^2 + \left\|W_l\right\|_F^2\right), \quad (9)$$
$$+ \lambda\left(\left\|P_lX_l - M_l\right\|_F^2 - N_l\left\|P_lX_l - \overline{M_l}\right\|_F^2\right)$$
$$s.t.\ e^T D_l = e^T, S_l \geq 0, diag(W_l) = 0$$

where $\left\|X_l^T - S_l^T D_l^T\right\|_{2,1} + \left\|S_l^T - X_l^T P_l^T\right\|_{2,1}$ is the $l_{2,1}$-norm based approximation error. Note that $D$, $P$, $S$ and $W$ are initialized to be random matrices with unit F-norm. Then, the above minimization can be alternated among the following steps:

*1) Fix the adaptive weight matrix W, update P, S, D:*

Given $W$, we can compute the analysis dictionary $P$, coding coefficients $S$ and synthesis dictionary $D$ from Eq.(9). We first show the optimization of $P$. By removing terms irrelevant to $P$, we have the following degenerated problem:

$$\min_P \Gamma(P)=\sum_{l=1}^{c}\left\|S_l^T - X_l^T P_l^T\right\|_{2,1} + \alpha\left(\left\|P_l\overline{X_l}\right\|_F^2 + \left\|P_l^T\right\|_{2,1}\right) \quad . \quad (10)$$
$$+ \beta\left\|P_lX_l - P_lX_lW_l\right\|_F^2 + \lambda\left(\left\|P_lX_l - M_l\right\|_F^2 - N_l\left\|P_lX_l - \overline{M_l}\right\|_F^2\right)$$

Based on the definition of $l_{2,1}$-norm [17][34][43-44], we have $\left\|S_l^T - X_l^T P_l^T\right\|_{2,1} = 2tr\left((S_l - P_lX_l)U_l(S_l^T - X_l^T P_l^T)\right)$ and $\left\|P_l^T\right\|_{2,1} = 2tr(P_lH_lP_l^T)$, where $H_l$ is a diagonal matrix with the $(i, i)$-th diagonal entries $H_{l,ii} = 1/\left[2\left\|\left(P_l^T\right)^i\right\|_2\right]$, $\left(P_l^T\right)^i$ is the $i$-th column vector of $P_l$ and $U_l$ is also a diagonal matrix with the $(i, i)$-th entries $U_{l,ii} = 1/\left[2\left\|\left(S_l^T - X_l^T P_l^T\right)^i\right\|_2\right]$. In reality, since $\left\|\left(P_l^T\right)^i\right\|_2$ may be equal to 0, we can use $2\left\|\left(P_l^T\right)^i\right\|_2 + \tau$ to approximate $2\left\|\left(P_l^T\right)^i\right\|_2$ under those cases. Similar argument exists for $\left\|\left(S_l^T - X_l^T P_l^T\right)^i\right\|_2$. Then, we can rewrite the formulation $\Gamma(P)$ as

$$\Gamma(P,U_l,H_l)=\sum_{l=1}^{c}2tr\left((S_l - P_lX_l)U_l(S_l^T - X_l^T P_l^T)\right)$$
$$+ \alpha\left(\left\|P_l\overline{X_l}\right\|_F^2 + 2tr(P_lH_lP_l^T)\right) + \beta tr(P_lX_lQ_lX_l^T P_l^T), \quad (11)$$
$$+ \lambda\left(\left\|P_lX_l - M_l\right\|_F^2 - N_l\left\|P_lX_l - \overline{M_l}\right\|_F^2\right)$$

where $Q_l = (I - W_l)(I - W_l)^T$. When the vectors $\left\|\left(S_l^T - X_l^T P_l^T\right)^i\right\|_2 \neq 0$ and $\left\|\left(P_l^T\right)^i\right\|_2 \neq 0$, by defining the derivatives and setting the derivative $\delta\Gamma(P,U_l,H_l)/\delta P = 0$, we can infer $P_l^{t+1}$ at the $(t+1)$-th iteration as follows:

$$P_l^{t+1} = \left(4S_l^t U_l^t X_l^T + 2\lambda M_l X_l^T - 2\lambda N_l\overline{M_l}X_l^T\right)\times\left(\Lambda^t + \Upsilon^t\right)^{-1}, \quad (12)$$

where $\Lambda^t = 4X_lU_l^t X_l^T + 2\alpha\overline{X_l}\overline{X_l}^T + 4\alpha H_l^t$, $\Upsilon^t = \beta X_lQ_lX_l^T + \beta X_lQ_l^T X_l^T + 2\lambda(N_l - 1)X_lX_l^T$, $\alpha, \beta$ and $\lambda$ are constant parameters.

After obtaining $P_l^{t+1}$, the mean matrices $M_l$ and $\overline{M_l}$ can be updated accordingly. We then describe the optimization of $S$. Similarly by removing the terms that are irrelevant to $S$, we can have the following reduced problem:

$$\min_{S,V_l,U_l} \Theta(S,V_l,U_l)=\sum_{l=1}^{c}\left\|X_l^T - S_l^T D_l^T\right\|_{2,1} + \left\|S_l^T - X_l^T P_l^T\right\|_{2,1}$$
$$= \sum_{l=1}^{c}2tr\left((X_l - D_lS_l)V_l(X_l^T - S_l^T D_l^T)\right) \quad , \quad (13)$$
$$+ 2tr\left((S_l - P_lD_l)U_l(S_l^T - X_l^T P_l^T)\right), \quad s.t.\ S_l \geq 0$$

where $V_l$ is also a diagonal matrix with the diagonal entries being $V_{l,ii} = 1 / \left[ 2 \left\| \left( X_l^T - S_l^T D_l^T \right)^i \right\|_2 \right]$ and $\left( X_l^T - S_l^T D_l^T \right)^i$ is the $i$-th row of $X_l^T - S_l^T D_l^T$. Note that the above equation holds when each $\left\| \left( X_l^T - S_l^T D_l^T \right)^i \right\|_2 \neq 0$ and $\left\| \left( S_l^T - X_l^T P_l^T \right)^i \right\|_2 \neq 0$.

Let $\psi_{ik}$ be the Lagrange multiplier for constraint $s_{l,ik} \geq 0$ [18][19] and $\Psi = [\psi_{ik}]$, we can deduce that the Lagrange function $\zeta$ to Eq.(13) can be formulated as

$$\zeta = 2tr\left( (X_l - D_l S_l) V_l \left( X_l^T - S_l^T D_l^T \right) \right) \\ + 2tr\left( (S_l - P_l D_l) U_l \left( S_l^T - X_l^T P_l^T \right) \right) + tr(\Psi S_l^T) \quad . \quad (14)$$

The partial derivatives of $\zeta$ w.r.t. variable $S_l$ is defined as

$$\partial \zeta / \partial S_l = -4 D_l^T X_l^T V + 4 D_l^T D_l S_l V_l + 4 S_l U_l - 4 P_l X_l U_l + \Psi . \quad (15)$$

Based on the KKT condition [18][35], i.e., $\psi_{ik} s_{l,ik} = 0$, we can obtain the following equation for $s_{l,ik}$:

$$-\left(4 D_l^T X_l^T V_l\right)_{ik} s_{l,ik} + \left(4 D_l^T D_l S_l V_l\right)_{ik} s_{l,ik} \\ + \left(4 S_l U_l\right)_{ik} s_{l,ik} - \left(4 P_l X_l U_l\right)_{ik} s_{l,ik} = 0 \quad , \quad (16)$$

which can lead to the following rules to update the element of the $i$-th row and $k$-th column of $S_l^{t+1}$:

$$s_{l,ik}^{t+1} \leftarrow s_{l,ik}^t \frac{\left( D_l^{tT} X_l^T V_l^t + P_l^{t+1} X_l U_l^t \right)_{ik}}{\left( D_l^{tT} D_l^t S_l^t V_l^t + S_l^t U_l^t \right)_{ik}} . \quad (17)$$

After $S$ is updated, we can show the optimization of $D$. By removing the terms that are irrelevant to $D$, the problem w.r.t. $D$ can be reformulated as follows:

$$\min_{D, V_l} \Omega(D, V_l) = \sum_{l=1}^{C} \left\| X_l^T - S_l^T D_l^T \right\|_{2,1} \\ = 2tr\left( (X_l - D_l S_l) V_l \left( X_l^T - S_l^T D_l^T \right) \right), \ s.t. \ e^T D_l = e^T \quad , \quad (18)$$

when each vector $\left\| \left( X_l^T - S_l^T D_l^T \right)^i \right\|_2 \neq 0$. By setting the derivative $\delta \Omega(D, V_l) / \delta D = 0$, we can update $D_l$ as

$$D_l^{t+1} = \left( X_l V_l^t S_l^{(t+1)T} \right) \left( S_l^{t+1} V_l^t S_l^{(t+1)T} + \tau I \right)^{-1} \text{ and } e^T D_l^{t+1} = e^T, \quad (19)$$

where $\tau$ is a small number to avoid the singularity and make the inverse computation stable and the operation $e^T D_l^{t+1} = e^T$ means that the atoms in $D_l^{t+1}$ are normalized.

*2) Fix P, S and D, update $H_l$, $U_l$ and $V_l$:*
With $P$, $S$, and $D$ computed, we can easily update the entries of the three diagonal matrices $H_l$, $M_l$ and $V_l$ by

$$H_l^{t+1} = diag\left( H_{l,ii}^{t+1} \right), H_{l,ii}^{t+1} = 1 / \left[ 2 \left\| \left( P_l^{(t+1)T} \right)^i \right\|_2 \right] \\ U_l^{t+1} = diag\left( U_{l,ii}^{t+1} \right), U_{l,ii}^{t+1} = 1 / \left[ 2 \left\| \left( S_l^{(t+1)T} - X_l^T P_l^{(t+1)T} \right)^i \right\|_2 \right] . \quad (20) \\ V_l^{t+1} = diag\left( V_{l,ii}^{t+1} \right), V_{l,ii}^{t+1} = 1 / \left[ 2 \left\| \left( X_l^T - S_l^{(t+1)T} D_l^{(t+1)T} \right)^i \right\|_2 \right]$$

*3) Fix P, update W*:
After the analysis dictionary $P$ is computed, we can update the adaptive weights $W$ by reformulating Eq.(8) as

$$W^* = \arg\min_W \Phi(W) = \sum_{l=1}^{c} \left\{ tr\left( (X_l - X_l W_l)(X_l^T - W_l^T X_l^T) \right) \\ + tr\left( (P_l X_l - P_l X_l W_l)(X_l^T P_l^T - W_l^T X_l^T P_l^T) \right) + tr(W_l W_l^T) \right\} . \quad (21) \\ s.t. \ diag(W_l) = 0$$

By setting derivative $\delta \Phi(W) / \delta W = 0$, we can update $W$ as

$$W_l^{t+1} = \left( X_l^T X_l + X_l^T P_l^{(t+1)T} P_l^{t+1} X_l + I \right)^{-1} \\ \times \left( X_l^T X_l + X_l^T P_l^{(t+1)T} P_l^{t+1} X_l \right) \quad , \quad (22)$$

and $\left[ W_l^{t+1} \right]_{ii} = 0$, where $\left[ W_l^{t+1} \right]_{ii}$ is the $(i, i)$-th diagonal entry of $W_l^{t+1}$. So, $\left[ W_l^{t+1} \right]_{ii} = 0$ means that all the diagonal entries are set to be 0 to avoid the trivial solution that $W_l^{t+1}$ is an identity matrix. For complete presentation, we summarize the optimization procedures of our RA-DPL in Table I. The learning algorithm iteratively optimizes each variable until the difference of the consecutive objective function values of Eq.(6) in adjacent iterations is less than $10^{-4}$ in the simulations. The diagonal matrices $H_l$, $U_l$ and $V_l$ are initialized to be identity matrices, similarly as the existing [11][17] that have proved that this way of initialization can generally perform well in most cases.

**Table I**: Robust Adaptive Projective Dictionary Pair Learning

**Input:** Training data matrix $X$, class label set $Y$, dictionary size $K$, parameters $\alpha$, $\beta$ and $\lambda$.
**Output:** $D = [D_1, \cdots D_l, \cdots D_c] \in \mathbb{R}^{n \times K}$, $S = [S_1, \cdots S_l, \cdots S_c] \in \mathbb{R}^{K \times N}$,

$$P = [P_1; \cdots P_l; \cdots P_c] \in \mathbb{R}^{K \times n}, W = \begin{bmatrix} W_1 & 0 & 0 \\ 0 & \cdots & 0 \\ 0 & 0 & W_c \end{bmatrix} \in \mathbb{R}^{N \times N}.$$

1: Initialize $P^{(0)}$, $S^{(0)}$, $W^{(0)}$ and $D^{(0)}$ as random matrices with unit F-norm; Initialize $H_l^{(0)}, U_l^{(0)}, V_l^{(0)}$ as identity matrices; $t \leftarrow 0$;
2: **while** *not converge* **do**
3:    **for** $l \leftarrow 1, 2, \cdots, c$ **do**
4:    Update the analysis dictionary $P_l^{(t+1)}$ by
$$P_l^{t+1} = \left( 4 S_l^t U_l^t X_l^T + 2\lambda M_l X_l^T - 2\lambda N_l \overline{M_l} X_l^T \right) \times \left( \Lambda^t + \Upsilon^t \right)^{-1},$$
where $\Lambda^t = 4 X_l U_l^t X_l^T + 2\alpha \overline{X_l} \overline{X_l}^T + 4\alpha H_l^t$,
$$\Upsilon^t = \beta X_l Q_l X_l^T + \beta X_l Q_l^T X_l^T + 2\lambda (N_l - 1) X_l X_l^T ;$$
5:    Update the sparse coefficients $S_l^{(t+1)}$ by
$$s_{l,ik}^{t+1} \leftarrow s_{l,ik}^t \frac{\left( D_l^{tT} X_l^T V_l^t + P_l^{t+1} X_l U_l^t \right)_{ik}}{\left( D_l^{tT} D_l^t S_l^t V_l^t + S_l^t U_l^t \right)_{ik}} ;$$
6:    Update the synthesis dictionary $D_l^{(t+1)}$ by
$$D_l^{t+1} = \left( X_l V_l^t S_l^{(t+1)T} \right) \left( S_l^{t+1} V_l^t S_l^{(t+1)T} + \tau I \right)^{-1} \text{ and } e^T D_l^{t+1} = e^T ;$$
7:    Update the diagonal matrices $H_l^{(t+1)}, U_l^{(t+1)}, V_l^{(t+1)}$ by
$$H_l^{t+1} = diag\left( H_{l,ii}^{t+1} \right), H_{l,ii}^{t+1} = 1 / \left[ 2 \left\| \left( P_l^{(t+1)T} \right)^i \right\|_2 \right]$$
$$U_l^{t+1} = diag\left( U_{l,ii}^{t+1} \right), U_{l,ii}^{t+1} = 1 / \left[ 2 \left\| \left( S_l^{(t+1)T} - X_l^T P_l^{(t+1)T} \right)^i \right\|_2 \right] ;$$
$$V_l^{t+1} = diag\left( V_{l,ii}^{t+1} \right), V_{l,ii}^{t+1} = 1 / \left[ 2 \left\| \left( X_l^T - S_l^{(t+1)T} D_l^{(t+1)T} \right)^i \right\|_2 \right]$$
8:    Update adaptive reconstruction weight matrix $W_l^{(t+1)}$ by
$$W_l^{t+1} = \left( X_l^T X_l + X_l^T P_l^{(t+1)T} P_l^{t+1} X_l + I \right)^{-1} \left( X_l^T X_l + X_l^T P_l^{(t+1)T} P_l^{t+1} X_l \right);$$
9: *end* **for**
10:   $t \leftarrow t + 1$; *end* **while**

*C. Convergence Analysis*

The problem of RA-DPL is solved alternately, so we would like to analyze its convergence. Note that our RA-DPL is an alternate convex search (ACS) algorithm [20-22], so we can have the following remarks [20-22] to assist the analysis.

**Theorem 1** [22]. If $B \in \mathbb{R}^{n \times m}$ is a bi-convex set, $f : B \to \mathbb{R}$ is bounded and the optimization of the variables in each iteration are solvable, the generated sequence $\{f(z_i)\}_{i \in t}$ ($z_i \in B$) by using the ACS algorithm will converge.

**Theorem 2** [22]. Let $X \subseteq \mathbb{R}^n, Y \subseteq \mathbb{R}^m$ be the closed set and let $f : X \times Y \to \mathbb{R}$ be continuous. Let the optimization of each variable in each iteration be solvable, then we can have:

(1) Suppose that the sequence $(z_i)_{i \in t}$ by ACS is contained within a compact set, the sequence will contain at least one

accumulation point.

(2) For each accumulation point $z^*$ of sequence $(z_i)_{i \in t}$, a) if the optimal solution of one variable with the others fixed in each iteration is unique, then all accumulation points will be the local optimal solutions and have the same function value; b) if the optimal solution of each variable is unique, then we have $\lim_{i \to \infty} \|z_{i+1} - z_i\| = 0$, and the accumulation points can form a compact continuum $C$.

Based on the Theorem 1 and Theorem 2, we can present three remarks on the convergence of our RA-DPL.

**Remark 1.** The generated sequence $\{f(D^i, S^i, P^i, W^i)\}_{i \in t}$ by our RA-DPL algorithm converges monotonically, given the diagonal matrices $H_t$, $U_t$ and $V_t$ in each iteration.

*Proof.* For our RA-DPL problem in Eq. (9), the variables $W$, $P$, $S$ and $D$ are main variables to be optimized. From the optimization procedures, if $W$ is fixed, variables $P$, $S$ and $D$ can be optimized alternately and can be treated as a single variable. If $P$, $S$ and $D$ are fixed, the variable $W$ can be optimized respectively as a single variable. As such, the problem in Eq. (9) is a bi-convex problem based on the combination $\{(D, P, S), (W)\}$. According to [22], the optimal solutions of $(D, P, S)$ and $W$ correspond to the iteration steps in ACS, and the problem has a general lower bound 0 due to the summarization of norms. Thus, based on Theorem 1, the sequence $\{f(D^i, S^i, P^i, W^i)\}_{i \in t}$ generated by our RA-DPL can converge monotonically.

**Remark 2.** The sequence of $\{D^i, S^i, P^i, W^i\}_{i \in t}$ generated by our RA-DPL algorithm has at least one accumulation point. All the accumulation points are the local optimal solutions of $f$ and moreover have the same function value.

*Proof.* Suppose $\|P_i^T\|_{2,1} \to \infty$, we have $f(D, S, P, W) \to \infty$. Thus, $\{D^i, S^i, P^i, W^i\}_{i \in t}$ is bounded in finite dimensional space, and the compact set condition in Theorem 2 (*Condition 1*) is met. Thus, the sequence has at least one accumulation point. By *Theorem 2 (Condition 2a)*, all the accumulation points are local optimal and have the same functional value.

**Remark 3.** Suppose $D$, $W$ and $P$ have unique solutions, sequence $\{D^i, S^i, P^i, W^i\}_{i \in t}$ generated by RA-DPL satisfies:

$$\lim_{i \to \infty} \|D^{i+1} - D^i\| + \|S^{i+1} - S^i\| + \|P^{i+1} - P^i\| + \|W^{i+1} - W^i\| = 0. \quad (23)$$

*Proof.* Based on *Remark 2*, the *Condition 1* and *2a* in the Theorem 2 are satisfied in RA-DPL, if we have the unique optimal solution of $D, P$, then we have the conclusion Eq. (23) based on the *Condition 2b* in *Theorem 2* [22]. Thus, it is easy to check that our RA-DPL is a reasonable approach.

### D. Time Complexity Analysis

We analyze the time complexity of our RA-DPL method. In the training phase, the variables $P_t, S_t, D_t, W_t$ and $H_t, U_t, V_t$ are undated alternately. In each Iteration, the time complexities of updating $P_t, S_t, D_t, W_t$ are $O(k_t N_t^2 + n N_t^2 + n^3 + k_t n^2)$, $O(k_t N_t^2 + k_t^2 n + k_t^3 + k_t^2 n + k_t n^2)$, $O(n N_t^2 + N_t n k_t + k_t^3 + k_t^2 N_t + k_t N_t^2 + n k_t^2)$, $O(N_t n^2 + N_t n k_t + N_t^2 n + N_t^3)$ and the complexities of updating $H_t, U_t, V_t$ are $O(k_t)$, $O(N_t n k_t)$ and $O(n k_t^2)$, respectively.

In the test phase, this classification scheme is very efficient. The computation of reconstruction error $\|y - D_i P_i y\|_2$ only has a complexity of $O(n k_t)$. Thus, the total complexity to classify the test set with $N_{test}$ samples is $O(N_{test} n k_t)$.

### E. Classification Approach

After convergence of our RA-DPL, the robust analysis sub-dictionary $P_k^*$ can be trained to produce small coefficients of data from the classes other than $k$, and it can only generate the significant coefficients for samples of class $k$. Meanwhile, the synthesis sub-dictionary $D_k^*$ by the robust reconstruction is also trained to reconstruct data of class $k$ from their coefficients $P_k^* X_k^*$, i.e., the residual $\|X_k^T - X_k^T P_k^{*T} D_k^{*T}\|_{2,1}$ will be potentially small similarly as [9]. On the other hand, since $D_k^*$ is not trained to reconstruct $X_i$ and $P_k^* X_i$ ($i \ne k$) is small, the residual $\|X_i^T - X_i^T P_k^{*T} D_k^{*T}\|_{2,1}$ will be much larger. In the testing phase, if a query sample $y$ is from class $k$, its projective sparse coding vector by the robust $P_k^*$ will be more likely to be significant, while its sparse coding vector by $P_i$ ($i \ne k$) tends to be small. Therefore, the reconstruction residual $\|y - D_k^* P_k^* y\|_2^2$ tends to be much smaller than residual $\|y - D_i^* P_i^* y\|_2^2, i \ne k$. As such, the class-specific reconstruction residual can be used to identify the class label of sample $y$. Thus, we naturally define the following classifier associated with our RA-DPL model similarly as [4][9]:

$$identity(y) = \arg\min_i \|y - D_i P_i y\|_2, \quad (24)$$

where $P$ is a robust analysis dictionary and is also a projection for the extraction of sparse coding coefficients.

## IV. EXPERIMENTAL RESULTS AND ANALYSIS

We mainly evaluate our RA-DPL for the data representation and classification. The performance of RA-DPL is mainly compared with those of related SRC [4], DLSI [8], KSVD [3], D-KSVD [6], LC-KSVD [1], FDDL [7], ADDL [11], DPL [9] and LRSDL [24]. Since DLSI and KSVD did not define an explicit classification model, we apply the same approach as SRC for DLSI and KSVD. In this study, five face databases (i.e., ORL [25], YaleB [26], UMIST [27], AR [28] and CMU PIE [29]), an object database (i.e., ETH80 [30]) and a scene database (i.e., the fifteen scene categories database [31]) are used for the evaluations. Note that these datasets are widely used to evaluate the DL methods [1-11]. Details of these datasets are shown in Table II, in which we report the number of samples, dimension and the number of subjects. In our simulations, the images of ORL, AR, YaleB, CMU PIE, UMIST and ETH80 are all resized into 32×32 pixels, thus each image can correspond to a data point in a 1024-D space. For classification, we randomly split each set into a training set and a test set. For fair comparison to other algorithms, the classification accuracy is averaged over 10 random splits of training and test samples to avoid the bias caused by randomness. We perform all simulations on a PC with Intel (R) Core (TM) i7-7700 CPU @ 3.6 GHz 8G.

TABLE II.
DESCRIPTIONS OF USED REAL-WORLD IMAGE DATABASES.

| Dataset Name | # Samples | # Dim | # Classes |
|---|---|---|---|
| ORL face | 400 | 1024 | 40 |
| YaleB face | 2414 | 504 | 38 |
| AR face | 2600 | 540 | 100 |
| CMU PIE face | 11554 | 1024 | 68 |
| UMIST face | 1012 | 1024 | 20 |
| ETH80 object | 3280 | 1024 | 80 |
| Fifteen scene categories | 4485 | 3000 | 15 |

### A. Convergence Analysis

We first analyze the convergence behavior by describing the objective function values. The ORL, AR, YaleB, CMU PIE, UMIST and ETH80 databases are used, and we select 5, 20, 20, 30, 10 and 6 images from each subject for the training set respectively, and set the dictionary size as the number of training samples. For AR and YaleB, we set the number of atoms corresponding to an average of 5 items per person. To be consistent with the following recognition simulations, we apply the random face features [1-3][13][19] for AR and YaleB, and the dimensions of extracted features are 540 and 504 respectively. The averaged results over 30 iterations are presented in Fig.2. We find that the objective function value of our RA-DPL is non-increasing in the iterations, and finally converges into a fixed value. The number of iterations in our RA-DPL is usually less than 20 in most cases.

### B. Parameter Selection Analysis

We present the parameter sensitivity analysis of RA-DPL in this study. Since the parameter selection issue still remains an open problem, we use a heuristic way to select the most important parameters. Note that our RA-DPL includes three parameters (i.e., $\alpha, \beta$ and $\lambda$), thus we aim to fix one of the parameters and explore the effects of other two on the test performance by using the grid search strategy. In this study, the YaleB face database is used as an example. We use random face features, i.e., each face image is projected onto a 504-D vector by a generated matrix from a zero-mean normal distribution, and each row of matrix is $l_2$ normalized. A half of face images is randomly chosen for training and the number of atoms is set to 760. For each pair of parameters, we average the results over 10 random splits of training and testing samples with varied parameters $\alpha, \beta$ and $\lambda$ from candidate set $\{10^{-5}, 5\times10^{-5}, \ldots, 10^{-1}, 5\times10^{-1}, 5, 5\times10^{1}, \ldots, 5\times10^{3}\}$. The results of the parameter selection are illustrated in Fig.3. We can find that: (1) RA-DPL can generally perform well in a wide range of the selections of parameters $\alpha$ and $\beta$ in each group, which means that our proposed RA-DPL is insensitive to $\alpha$ and $\beta$; (2) a large value of $\lambda$ tends to decrease the recognition results, which may be because the locality preservation term associated with parameter $\lambda$ has a large effect on the performance.

In addition to the above visual parameter analysis, we also investigate the effects of the three components in the objective function of our RA-DPL on the result by setting $\alpha = 0$, $\beta = 0$ and $\lambda = 0$, respectively. In this simulation, four image databases, i.e., CMU PIE, UMIST, Fifteen scene and ETH80, are evaluated. We respectively train on 30, 5, 40 and 6 images from each subject for CMU PIE, UMIST, Fifteen scene and ETH80, and the remaining images are used for testing. Moreover, the number of atoms is set to 2040, 100, 450 and 480, respectively. The results are shown in the Table III. We can clearly find that when $\alpha = 0$ (i.e., $\|P_l \overline{X_l}\|_F^2 + \|P_l^T\|_{2,1}$ is removed), the classification performance of our algorithm is decreased significantly. When $\beta = 0$ (i.e., the adaptive weight learning is removed), the performance is also inferior to the full model, implying that preserving the neighborhood relationship within each subject class is indispensable. When $\lambda = 0$ (i.e., the discriminating function on the coding coefficients is removed), the performance of our algorithm is also decreased. By the above parameter analysis, we can easily conclude that the any component in the objective function of our method are all important for improving the performance of our algorithm.

Additionally, we also carefully investigate the hyperparameter selection issues of the other evaluated competitors using the grid search strategy or linear search strategy from the same candidate set for fair comparison. For each method and hyperparameter, we repeat the results over 10 random splits of training/test images, and the averaged image recognition accuracies are reported for the fair comparison.

TABLE III
COMPARISON OF RECOGNITION RESULTS ON CMU PIE, UMIST, FIFTEEN SCENES AND ETH80 UNDER DIFFERENT PARAMETERS.

| Methods \ Datasets | CMU PIE | UMIST | Fifteen scenes | ETH80 |
|---|---|---|---|---|
| RA-DPL with $\alpha=0$, $\beta \neq 0$, $\lambda \neq 0$ | 93.1% | 86.0% | 92.5% | 95.0% |
| RA-DPL with $\alpha \neq 0$, $\beta=0$, $\lambda \neq 0$ | 93.6% | 87.5% | 96.0% | 96.7% |
| RA-DPL with $\alpha \neq 0$, $\beta \neq 0$, $\lambda=0$ | 93.2% | 88.6% | 95.9% | 96.5% |
| RA-DPL with $\alpha \neq 0$, $\beta \neq 0$, $\lambda \neq 0$ | **94.2%** | **92.1%** | **96.2%** | **98.1%** |

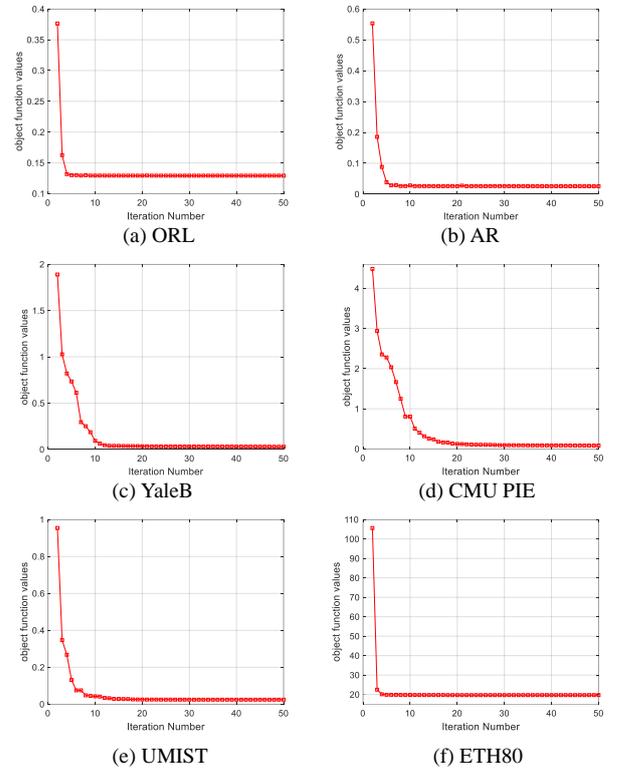

(a) ORL  (b) AR  (c) YaleB  (d) CMU PIE  (e) UMIST  (f) ETH80

**Fig.2:** Convergence behavior of RA-DPL on the evaluated databases, where the *x*-axis is the number of iterations and the *y*-axis represents the objective function values.

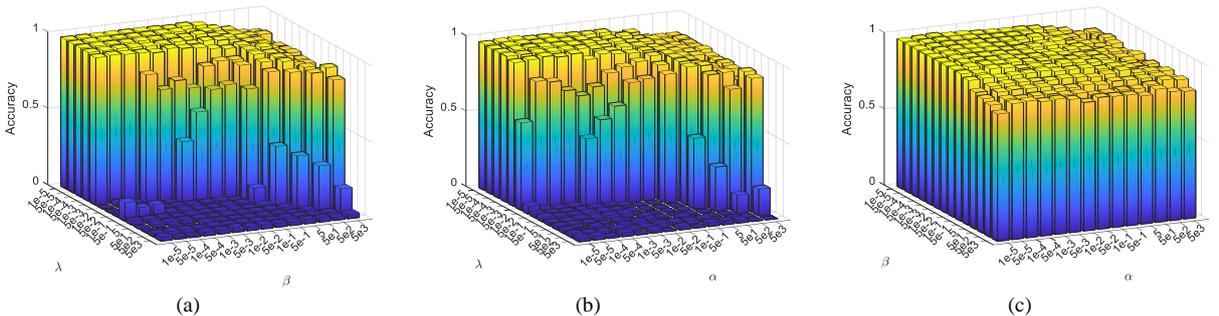

(a)  (b)  (c)

**Fig.3:** Parameter sensitivity of RA-DPL on YaleB face database, where (a) the effects of tuning $\lambda$ and $\beta$ on the performance by fixing $\alpha=0.00005$; (b) the effects of tuning $\lambda$ and $\alpha$ on the performance by fixing $\beta=0.0005$; (c) the effects of tuning $\alpha$ and $\beta$ on the performance by fixing $\lambda=0.00005$.

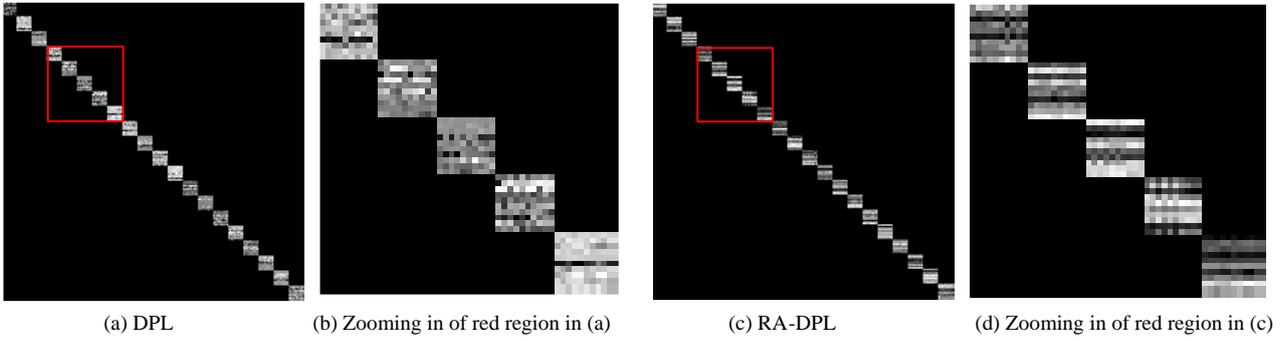

(a) DPL    (b) Zooming in of red region in (a)    (c) RA-DPL    (d) Zooming in of red region in (c)
**Fig.4**: Visualization of the computed coding coefficients *S* over training data by DPL and our RA-DPL in the training phase.

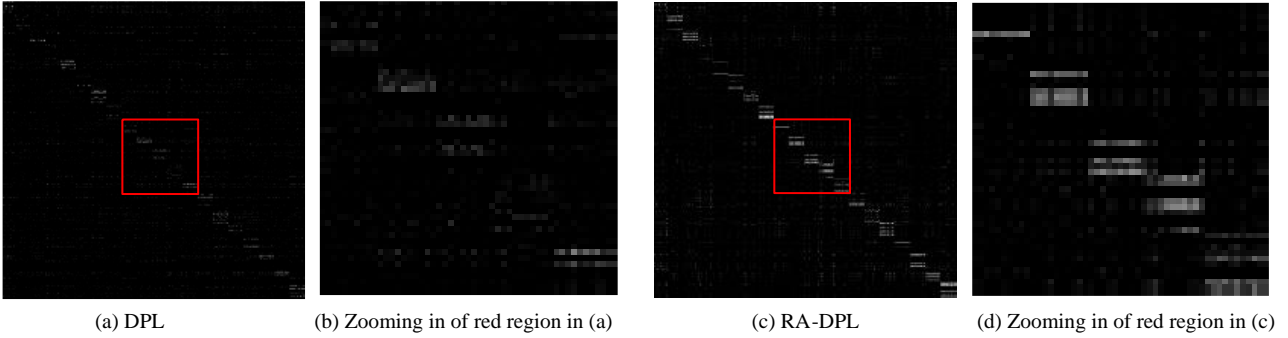

(a) DPL    (b) Zooming in of red region in (a)    (c) RA-DPL    (d) Zooming in of red region in (c)
**Fig.5**: Visualization of the approximated coefficients *PY* by embedding test data *Y* onto the projection *P* of DPL and our RA-DPL in testing phase.

More importantly, the reported results of each method are all based on the best choice of tuned hyperparameters.

### C. Exploratory Data Analysis by Visualization

We present the exploratory data analysis results by visualizing the coefficients, reconstruction error and reconstruction weights. UMIST face database is used and each person is shown in a range of pose from profile to fontal views. We select 10 images per person for training, 10 images from each person to form the test set *Y* for clear observation. The number of atoms is set to an average of 10 items per person.

**Visualization of sparse coefficients.** We first illustrate the sparse coefficients (*S*) of both DPL and RA-DPL from the training process in Fig.4, where the right figure is the zooming in of the red rectangle in the left figure. We can find that: (1) the sparse coefficients of DPL and RA-DPL over various classes are strictly block-diagonal due to the dictionary pair learning scheme; (2) the coding coefficients of RA-DPL are more sparse than those of DPL due to the sparse $l_{2,1}$-norm constraint on the analysis sub-dictionary. In addition, we also visualize the coefficients *PY* of DPL and RA-DPL over the test data *Y* from the testing phase in Fig.5. We can easily find that the block-diagonal structures of the computed coefficients by RA-DPL over various subjects in *Y* are clearer than those of DPL, and the connectivity in *PY* of our RA-DPL is better than that of DPL.

**Visualization of the reconstruction error on test data *Y*.** Since we used the $l_{2,1}$-norm to minimize the reconstruction error as much as possible, we would like to illustrate some quantitative evaluation results. More specifically, we apply the pre-learned dictionaries *D* and *P* to decompose *Y* into the recovered component *DPY* and an error part $Y - DPY$, and compute the quantitative reconstruction error $\|Y - DPY\|_F$ and peak signal-to-noise (PSNR) over recovered data *DPY*. We visualize the decomposition process of RA-DPL and DPL in Fig.6, where we also show the reconstruction error $\|Y - DPY\|_F$ and PSNR. From Fig.6, we can observe that the recovered data by RA-DPL is more accurate than that of DPL, since the reconstruction error by our RA-DPL is less than that of recent DPL, and the delivered PSNR value by our RA-DPL is also larger than that of DPL.

### D. Application to Image Recognition

We evaluate each method for representing and recognizing three kinds of images, i.e., face image databases (i.e., YaleB, AR, CMU PIE and UMIST), ETH80 object image database, and the fifteen nature scene categories database. Note that

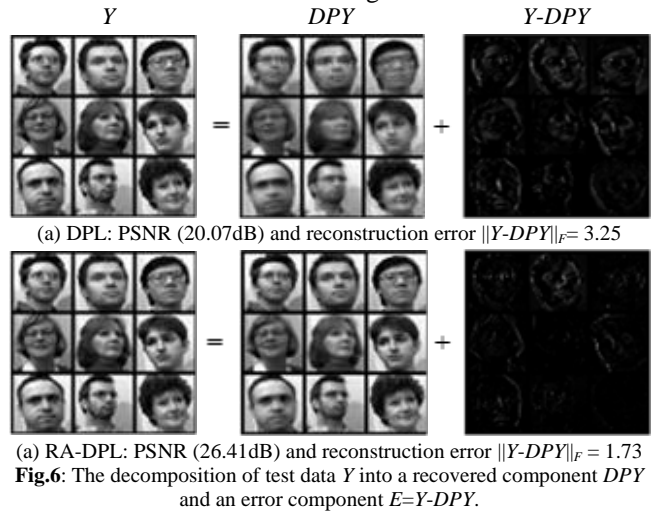

(a) DPL: PSNR (20.07dB) and reconstruction error $\|Y-DPY\|_F$= 3.25

(a) RA-DPL: PSNR (26.41dB) and reconstruction error $\|Y-DPY\|_F$ = 1.73
**Fig.6**: The decomposition of test data *Y* into a recovered component *DPY* and an error component *E=Y-DPY*.

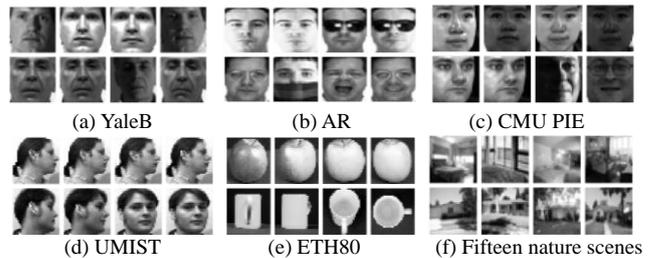

(a) YaleB    (b) AR    (c) CMU PIE
(d) UMIST    (e) ETH80    (f) Fifteen nature scenes
**Fig.7**: Sample images of the evaluated real image databases.

some examples of these databases are shown in Fig.7. The recognition results of RA-DPL are mainly compared with those of SRC, DLSI, KSVD, D-KSVD, LC-KSVD, COPAR,

FDDL, DPL, LRSDL and ADDL. For each algorithm, the model parameters are carefully chosen for fair comparison.

**Face Recognition on YaleB database.** This database has 2414 images of 38 people. Each person has 63 face images taken during two sessions. In this study, we use the random face features [1-3][13][19] and the dimension is set to 504. We strictly follow the settings in [9] for this study, i.e., half of the images per class are randomly selected for training and the rest is used for testing. The dictionary contains 570 atoms, corresponding to an average of 15 items per person. For each evaluated method, we repeat the experiments over 10 random splits of training and testing face images, and the recognition accuracy is reported as the average of different runs in this study for the fair comparison. In this study, $\alpha=0.0001$, $\beta=0.005$ and $\lambda=0.0001$ are used in our RA-DPL. The averaged recognition results are shown in Table IV, where the results of other compared methods are directly adopted from [9]. We can observe from the results that our RA-DPL outperforms its competitors under the same setting by achieving higher accuracies for this database.

TABLE IV
RECOGNITION RESULTS USING RANDOM FACE FEATURES ON YALEB.

| Evaluated Methods | Mean $\pm$ Std(%) |
|---|---|
| SRC(all train. sample) | 96.5 $\pm$ 0.85 |
| K-SVD(15 items) | 93.1 $\pm$ 0.85 |
| DKSVD(15 items) | 94.1 $\pm$ 0.80 |
| LC-KSVD1(15 items) | 94.5 $\pm$ 0.81 |
| LC-KSVD2(15 items) | 95.0 $\pm$ 0.79 |
| DLSI (15 items) | 97.0 $\pm$ 0.77 |
| COPAR (15 items) | 96.9 $\pm$ 0.72 |
| FDDL(15 items) | 96.7 $\pm$ 0.69 |
| DPL(15 items) | 97.5 $\pm$ 0.64 |
| LRSDL(15 items) | 97.3 $\pm$ 0.65 |
| ADDL(15 items) | 97.0 $\pm$ 0.62 |
| **Our RA-DPL(15 items)** | **97.8 $\pm$ 0.59** |

**Face Recognition on AR database.** This face database contains more than 4000 images from 126 people [11][28], and each person has 26 images taken during two sessions. Following the common evaluation procedures [1-3][12-14], the face set that contains 2600 images of 50 males and 50 females is employed. We also follow [1-3][13][19] to use random face features with the dimensionality being 540. We choose 20 images per person randomly for training and test on the rest. The dictionary has 500 atoms, corresponding to an average of 5 items per class. $\alpha=5\times10^{-5}$, $\beta=1$ and $\lambda=0.01$ are set for our RA-DPL. The results are shown in Table V, where the results of compared methods are adopted from [1] [11] directly. We find that RA-DPL obtains the enhanced results than its competing methods under the same setting.

TABLE V.
RECOGNITION RESULTS USING RANDOM FACE FEATURES ON AR.

| Evaluated Methods | Mean $\pm$ Std(%) |
|---|---|
| SRC (5 items, 20 labels) | 66.5 $\pm$ 3.89 |
| KSVD(5 items, 20 labels) | 86.5 $\pm$ 2.94 |
| DKSVD(5 items, 20 labels) | 88.8 $\pm$ 2.57 |
| LC-KSVD1(5 items, 20 labels) | 92.5 $\pm$ 2.04 |
| LC-KSVD2(5 items, 20 labels) | 93.7 $\pm$ 1.98 |
| DLSI(5 items, 20 labels) | 93.1 $\pm$ 1.77 |
| FDDL(5 items, 20 labels) | 95.6 $\pm$ 1.68 |
| DPL(5 items, 20 labels) | 95.8 $\pm$ 0.89 |
| LRSDL(5 items, 20 labels) | 96.8 $\pm$ 0.90 |
| ADDL(5 items, 20 labels) | 97.0 $\pm$ 0.96 |
| **Our RA-DPL(5 items, 20 labels)** | **97.7 $\pm$ 0.44** |

**Face Recognition on CMU PIE database.** CMU PIE face database contains 68 persons with 41368 face images as a whole. Follow the common procedures in [2][29], 170 near frontal images per person are employed for simulations. This face subset consists of five near frontal pose (C05, C07, C09, and C29) and all images have different illuminations, lighting and expression. We also use random face features as [4][17] and set the dimension to 256. For recognition, we train on 20, 30, and 40 images person and test on the rest, and set the number of dictionary atoms to the number of training images. $\alpha=10^{-5}$, $\beta=0.005$ and $\lambda=5\times10^{-5}$ are set in RA-DPL. The averaged results are described in Table VI, from which we can see that: (1) the recognition accuracy increases as the training number increases; (2) our RA-DPL is superior to its competitors in most cases, and the main reason for the improvement by RA-DPL can be attributed to keeping the local neighborhood information and its robust adaptive dictionary learning pair scheme.

TABLE VI
RECOGNITION RESULTS USING RANDOM FACE FEATURES ON CMU PIE.

| Evaluated Methods | 20 Mean $\pm$ Std(%) | 30 Mean $\pm$ Std(%) | 40 Mean $\pm$ Std(%) |
|---|---|---|---|
| SRC | 77.4 $\pm$ 1.55 | 82.6 $\pm$ 1.75 | 83.5 $\pm$ 3.54 |
| KSVD | 78.9 $\pm$ 1.63 | 83.0 $\pm$ 1.72 | 84.3 $\pm$ 3.33 |
| D-KSVD | 80.2 $\pm$ 1.42 | 83.5 $\pm$ 1.51 | 85.9 $\pm$ 3.15 |
| LC-KSVD1 | 81.3 $\pm$ 1.22 | 85.0 $\pm$ 1.38 | 87.1 $\pm$ 2.83 |
| LC-KSVD2 | 81.5 $\pm$ 1.11 | 85.9 $\pm$ 1.31 | 87.2 $\pm$ 2.77 |
| DLSI | 78.3 $\pm$ 0.89 | 84.5 $\pm$ 1.08 | 89.1 $\pm$ 2.27 |
| COPAR | 86.1 $\pm$ 0.99 | 89.1 $\pm$ 1.12 | 90.9 $\pm$ 2.39 |
| FDDL | 84.7 $\pm$ 0.96 | 89.5 $\pm$ 1.06 | 91.2 $\pm$ 1.96 |
| DPL | 86.5 $\pm$ 0.85 | 89.4 $\pm$ 1.02 | 90.3 $\pm$ 1.81 |
| LRSDL | 87.1 $\pm$ 0.81 | 89.5 $\pm$ 1.05 | 91.2 $\pm$ 1.94 |
| ADDL | 87.0 $\pm$ 0.78 | 89.6 $\pm$ 0.95 | 91.0 $\pm$ 1.63 |
| **Our RA-DPL** | **91.9 $\pm$ 0.40** | **94.2 $\pm$ 0.55** | **95.0 $\pm$ 1.26** |

**Face Recognition on UMIST database.** This database contains 1012 images of 20 individuals, and each individual is shown in a range of pose from profile to fontal views [27]. In this simulation, we randomly select 5 images per class for training and use other images for testing. The number of dictionary atoms is set to be the number of training samples. In this study, we normalize each sample to be unit $l_2$-norm. $\alpha=0.005$, $\beta=0.05$ and $\lambda=5\times10^{-5}$ are applied in our RA-DPL. The averaged results are shown in Table VII. We can find that our RA-DPL can obtain the enhanced results compared with other related algorithms. In addition, we also evaluate RA-DPL by using a smaller dictionary corresponding to 2 items per person. Once again, we can see that our RA-DPL can outperform other competitors for face recognition.

TABLE VII.
RECOGNITION RESULTS ON THE UMIST DATABASE

| Evaluated Methods | Mean $\pm$ Std(%) |
|---|---|
| SRC (5 items, 5 labels) | 87.4 $\pm$ 2.44 |
| KSVD(5 items, 5 labels) | 87.7 $\pm$ 2.49 |
| DKSVD(5 items, 5 labels) | 87.2 $\pm$ 2.13 |
| LC-KSVD1(5 items, 5 labels) | 87.8 $\pm$ 2.68 |
| LC-KSVD2(5 items, 5 labels) | 88.6 $\pm$ 1.95 |
| DLSI(5 items, 5 labels) | 87.1 $\pm$ 2.14 |
| FDDL(5 items, 5 labels) | 87.5 $\pm$ 1.64 |
| DPL(5 items, 5 labels) | 88.9 $\pm$ 1.62 |
| LRSDL(5 items, 5 labels) | 90.4 $\pm$ 2.31 |
| ADDL(5 items, 5 labels) | 90.9 $\pm$ 1.73 |
| **Our RA-DPL(2 items, 5 labels)** | **91.5 $\pm$ 1.48** |
| **Our RA-DPL(5 items, 5 labels)** | **92.1 $\pm$ 1.51** |

**Scene Recognition on fifteen categories database.** This database includes fifteen scenes, i.e., suburb, open country, mountain, coast, highway, forest, store, kitchen, industrial, office, living room, tall building, bedroom, street and inside city [31]. Each scene class has 200 to 400 images, and each image has about $250\times300$ pixels. Following [1], the spatial pyramid features by using a four-level spatial pyramid and a SIFT-descriptor codebook with size 200 are computed for simulations. The final spatial pyramid features are reduced to 3000 by PCA. Following the common settings in [1][5], we select 40 samples per class for training and test on the rest. The dictionary size is set to 450, corresponding to an average of 30 items over each class. $\alpha=5\times10^{-5}$, $\beta=5\times10^{-5}$ and $\lambda=5\times10^{-5}$ are used in our RA-DPL method.

We show the averaged recognition results in Table VIII, where we directly adopt the results of the other compared methods from [1][5]. We can find that our RA-DPL obtains higher accuracies than other models under the same setting. In addition, we also evaluate the recognition accuracy rates for individual scenes and Fig.8 shows some examples with the accuracies of each individual, from which we find that most of the confusion occurs between the indoor classes, for instance coast, mountain, and open country.

TABLE VIII.
RECOGNITION RESULTS USING SPATIAL FEATURES ON THE FIFTEEN SCENE CATEGORY DATABASE

| Evaluated Methods | Mean $\pm$ Std(%) |
|---|---|
| SRC (all train. sample) | 92.63 $\pm$ 1.52 |
| KSVD(30 items, 40 labels) | 85.62 $\pm$ 1.45 |
| DKSVD(30 items, 40 labels) | 87.32 $\pm$ 0.95 |
| LC-KSVD1(30 items, 40 labels) | 89.70 $\pm$ 1.12 |
| LC-KSVD2(30 items, 40 labels) | 91.60 $\pm$ 1.10 |
| DLSI (30 items, 40 labels) | 91.80 $\pm$ 1.25 |
| FDDL (30 items, 40 labels) | 92.16 $\pm$ 0.92 |
| DPL (30 items, 40 labels) | 95.08 $\pm$ 0.86 |
| LRSDL(30 items, 40 labels) | 95.14 $\pm$ 0.80 |
| ADDL(30 items, 40 labels) | 95.47 $\pm$ 0.77 |
| **Our RA-DPL(30 items, 40 labels)** | **96.20 $\pm$ 0.67** |

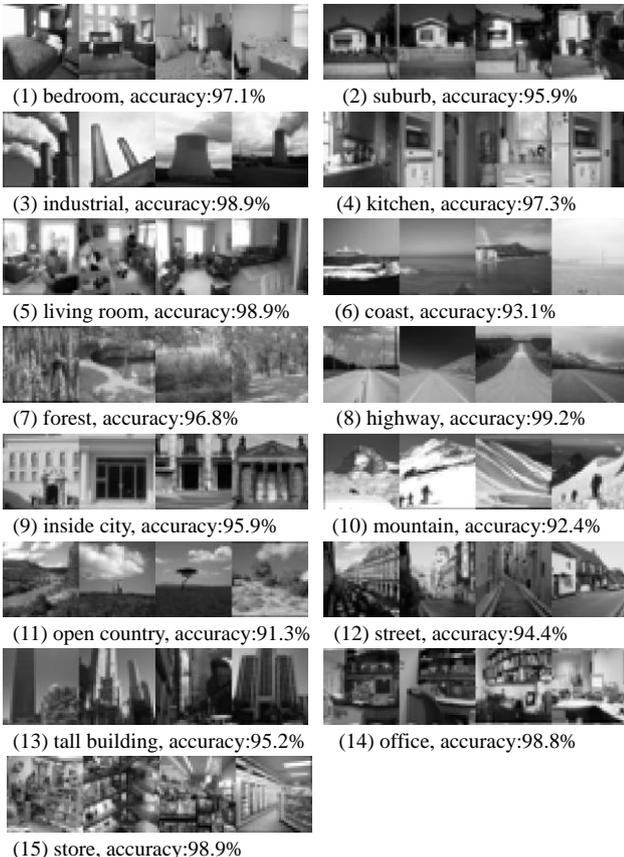

(1) bedroom, accuracy:97.1%   (2) suburb, accuracy:95.9%
(3) industrial, accuracy:98.9%   (4) kitchen, accuracy:97.3%
(5) living room, accuracy:98.9%   (6) coast, accuracy:93.1%
(7) forest, accuracy:96.8%   (8) highway, accuracy:99.2%
(9) inside city, accuracy:95.9%   (10) mountain, accuracy:92.4%
(11) open country, accuracy:91.3%   (12) street, accuracy:94.4%
(13) tall building, accuracy:95.2%   (14) office, accuracy:98.8%
(15) store, accuracy:98.9%

**Fig.8:** Image examples from the individual classes of fifteen nature scene categories database.

TABLE IX.
RECOGNITION RESULTS ON THE ETH80 OBJECT DATABASE

| Evaluated Methods | Mean $\pm$ Std (%) |
|---|---|
| SRC (6 items, 6 labels) | 89.6 $\pm$ 0.81 |
| KSVD(6 items, 6 labels) | 91.2 $\pm$ 0.79 |
| DKSVD(6 items, 6 labels) | 91.2 $\pm$ 0.42 |
| LC-KSVD1(6 items, 6 labels) | 90.7 $\pm$ 0.77 |
| LC-KSVD2(6 items, 6 labels) | 91.5 $\pm$ 0.85 |
| DLSI(6 items, 6 labels) | 92.7 $\pm$ 0.91 |
| FDDL(6 items, 6 labels) | 93.2 $\pm$ 0.35 |
| DPL(6 items, 6 labels) | 97.7 $\pm$ 0.22 |
| ADDL(6 items, 6 labels) | 97.9 $\pm$ 0.20 |
| LRSDL(6 items, 6 labels) | 97.7 $\pm$ 0.21 |
| **Our RA-DPL(6 items, 6 labels)** | **98.1 $\pm$ 0.14** |

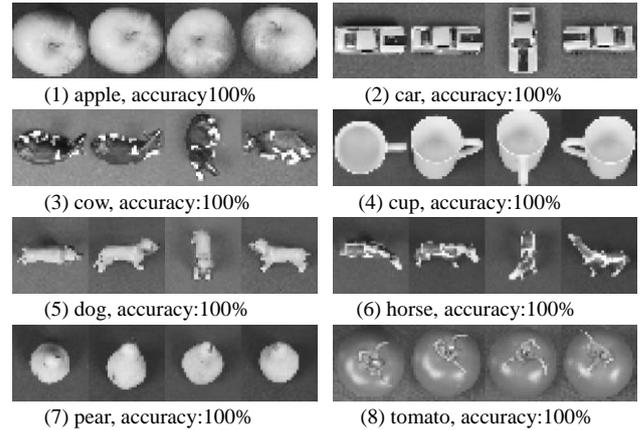

(1) apple, accuracy:100%   (2) car, accuracy:100%
(3) cow, accuracy:100%   (4) cup, accuracy:100%
(5) dog, accuracy:100%   (6) horse, accuracy:100%
(7) pear, accuracy:100%   (8) tomato, accuracy:100%

**Fig.9:** Image examples from the classes with highest accuracy rates from the ETH80 object database.

**Object Recognition on ETH80 database.** ETH80 object database has totally 3280 images of 80 subcategories from 8 big categories [30]. That is, it contains 8 big categories, including apple, car, cow, cup, dog, horse, pear and tomato. In each big category, 10 subcategories are included, each of which contains 41 images from different viewpoints. In this study, we follow [11] to perform dictionary learning based on discriminant features [33]. We select 6 images from each class for training and test on the rest. $\alpha=10$, $\beta=0.001$ and $\lambda=0.5$ are used in our RA-DPL. We show the averaged results in Table IX, from which we find that our RA-DPL achieves better performance than the other methods. ADDL also obtains promising results. In addition, we also evaluate the recognition rates for individual classes and show some image examples in the 8 object classes having 100 percent recognition accuracy rate in Fig.9.

### E. Image Recognition on Deep Convolutional Features

We investigate the image recognition tasks against the deep convolutional features [36][38][42]. For the consideration of efficiency, we use deep features as a preprocessing step to reduce the dimensionality from 1024 to 800. The used deep feature learning framework has two convolution and max pooling layers. Specifically, the first convolution layer uses $5\times5$ convolution kernel to handle each image to produce 16 feature maps of dimension $28\times28$, and the first pooling layer uses a $2\times2$ kernel and the stride length is 2 pixels, so it can output 16 feature map of dimension $14\times14$. The second convolution layer uses $5\times5$ kernel to process each image to output 32 feature map of dimension $10\times10$, and the second max pooling layer uses the $2\times2$ kernel and sets the stride length to 2 pixels, therefore 32 feature map of dimension $5\times5$ can be obtained. Two real face databases, i.e., AR and CMU PIE, are evaluated in this study.

TABLE X
RECOGNITION RESULTS USING CONVOLUTION FEATURES ON AR.

| Evaluated Methods | 5 Max acc | 5 Mean ± Std(%) | 10 Max acc | 10 Mean ± Std(%) |
|---|---|---|---|---|
| SRC | (all train. sample) | | 75.67% | 71.58 ± 0.78 |
| KSVD | 78.89% | 76.24 ± 2.19 | 80.82% | 77.03 ± 0.78 |
| D-KSVD | 80.87% | 78.65 ± 1.82 | 84.30% | 81.56 ± 0.78 |
| LC-KSVD1 | 84.50% | 82.53 ± 1.55 | 86.67% | 84.17 ± 0.78 |
| LC-KSVD2 | 84.83% | 83.16 ± 1.64 | 86.93% | 84.73 ± 0.78 |
| DLSI | 87.17% | 86.10 ± 1.96 | 89.83% | 87.05 ± 0.78 |
| COPAR | 88.67% | 86.46 ± 1.87 | 90.50% | 89.50 ± 0.78 |
| FDDL | 82.00% | 78.00 ± 2.73 | 89.50% | 84.34 ± 3.14 |
| DPL | 93.83% | 92.28 ± 1.34 | 94.00% | 92.28 ± 1.21 |
| LRSDL | 90.33% | 89.24 ± 1.28 | 94.00% | 93.23 ± 1.25 |
| ADDL | 93.80% | 92.33 ± 1.09 | 92.80% | 91.70 ± 0.86 |
| **RA-DPL** | **95.33%** | **93.45 ± 1.21** | **96.00%** | **94.60 ± 1.05** |

TABLE XI.
RECOGNITION RESULTS USING CONVOLUTION FEATURES ON CMU PIE.

| Evaluated Methods | 20 Mean ± Std(%) | 30 Mean ± Std(%) | 40 Mean ± Std(%) |
|---|---|---|---|
| SRC | 64.30 ± 1.12 | 68.56 ± 1.48 | 70.25 ± 2.06 |
| KSVD | 75.62 ± 1.02 | 77.23 ± 1.18 | 79.15 ± 1.76 |
| D-KSVD | 79.18 ± 0.83 | 82.26 ± 1.21 | 85.28 ± 1.58 |
| LC-KSVD1 | 84.56 ± 0.68 | 88.92 ± 0.97 | 90.52 ± 1.27 |
| LC-KSVD2 | 85.10 ± 0.63 | 89.25 ± 0.84 | 91.17 ± 1.08 |
| DLSI | 83.71 ± 0.59 | 89.74 ± 0.88 | 92.99 ± 1.02 |
| COPAR | 81.05 ± 0.61 | 87.82 ± 0.89 | 91.63 ± 1.09 |
| FDDL | 78.81 ± 0.69 | 83.28 ± 0.93 | 86.40 ± 1.48 |
| DPL | 88.05 ± 0.35 | 91.97 ± 0.56 | 93.75 ± 0.88 |
| LRSDL | 89.59 ± 0.32 | 93.67 ± 0.48 | 94.64 ± 0.75 |
| ADDL | 85.00 ± 0.30 | 91.18 ± 0.58 | 93.53 ± 0.84 |
| **RA-DPL** | **91.60 ± 0.24** | **94.49 ± 0.36** | **95.67 ± 0.53** |

**Results on AR face database.** In this simulation, we also randomly choose 5 and 10 images per person for training and use the rest for testing. The dictionary contains 500 and 1000 atoms, corresponding to an average of 5 and 10 items each class. $\alpha=5\times 10^{-5}$, $\beta=0.05$ and $\lambda=0.005$ are used in our RA-DPL. We report the maximum accuracy (Max acc) and averaged accuracy (Mean acc) over different runs to be the final recognition results that are shown in Table X. From the results, we can find that: (1) our RA-DPL is superior to its competing methods in most cases; (2) a large dictionary with more atoms can produce better recognition results.

**Results on CMU PIE database.** In this study, we choose 20, 30 and 40 mages per person for training and use the rest for testing. We set the number of atoms to the number of training samples. $\alpha=5\times 10^{-4}$, $\beta=0.5$ and $\lambda=5\times 10^{-4}$ are used in RA-DPL. We report the averaged results over different runs in Table XI. We see that: (1) the increasing number of training samples improves the performance of each method, since the labeled training data can provide the supervision information to improve the representation and classification powers potentially. Moreover, more supervision information are beneficial to higher accuracies as the number of training data is increased; (2) RA-DPL is superior to its competitors. LRSDL, DPL and ADDL can also work well by delivering better results than other remaining methods in most cases.

*F. Visual and Quantitative Investigation of Dictionaries*

We mainly evaluate the performance of the learned dictionary $D$ of RA-DPL, and show the comparison results to several related DL methods. We firstly visualize the dictionary for observation and then show the quantitative recognition results against varying dictionary sizes.

**Visualization of dictionary atoms.** We mainly visualize the learned dictionary $D$ of four structured DL methods and the UMIST face database is employed. We randomly select 5 images per class for training and use the other images for testing. The dictionary contains 100 items, corresponding to an average of 5 items each class. For a fair comparison of learned dictionary $D$ by each method, the dictionaries are initialized to be random matrices with unit F-norm and the number of iterations is set to 20. The visualization results of the learned dictionaries by structured DPL, ADDL, DLSI and our RA-DPL are illustrated in Fig.10. To measure the similarity between each class of dictionaries, we calculate the absolute value of the correlation coefficient between the dictionary atoms. Then, the resulting similarity matrix of each method is visualized using heat map. The visualization results of the heat map by each method are shown in Fig.11. From the heat maps, we can easily find that the similarities between each class of dictionaries by our RA-DPL are more accurate than those of DPL, ADDL and DLSI, because the heat maps of our RA-DPL contains less wrong inter-class connections and the connectivity within the sub-dictionaries of each class is also better than other methods. That is, the discriminating ability of the structured dictionary $D$ by our RA-DPL will be stronger than those of other algorithms.

**Quantitative recognition evaluation results by varying dictionary sizes.** Two face databases, i.e., MIT CBCL and CMU PIE, and the fifteen nature scene categories database are evaluated as the examples. The compared methods are LC-KSVD, DLSI, DPL, ADDL and FDDL. For CMU PIE, we still use random face features of dimension 256, choose 30 samples per class for training and evaluate each method with varying sizes $K$ of dictionary, i.e., $K$=340, 680, 1020, 1360, 1700 and 2040 in Fig.11a. For the fifteen nature scene database, we follow [1][5] to choose 100 samples per class for training and evaluate each method with varying sizes $K$ of the dictionary, i.e., $K$=75, 150, 225, 300, 375 and 450 in Fig.11b. For the MIT CBCL face database, it contains 3240 face images of 10 persons [37], i.e., 324 images per person rendered from 3D head models. We normalize each image data to have unit $l_2$-norm. We choose 6 samples per class as

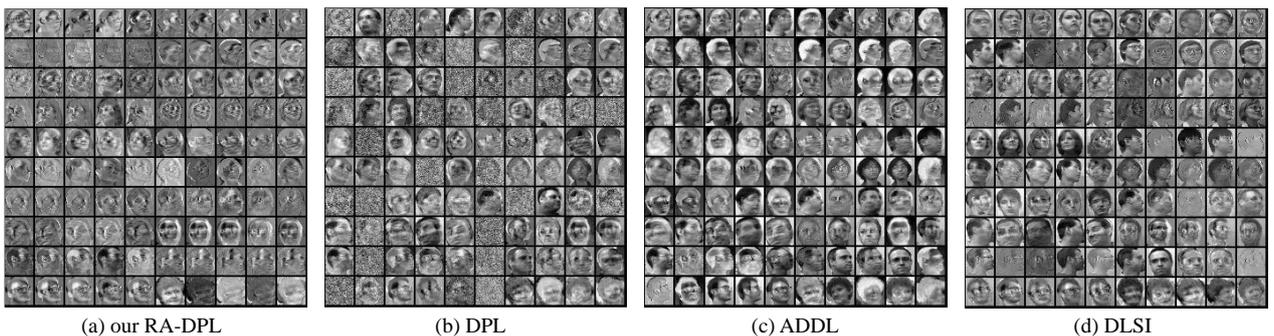

(a) our RA-DPL     (b) DPL     (c) ADDL     (d) DLSI
**Fig.10:** Visualization of the learned dictionary $D$ of each structured DL algorithm on the UMIST face database.

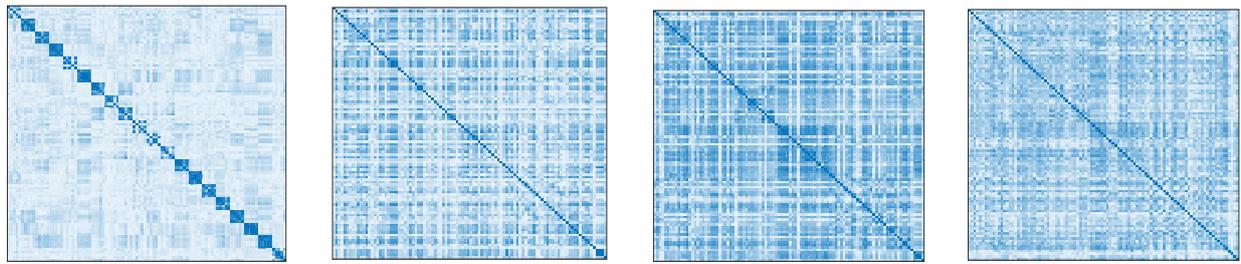

(a) our RA-DPL    (b) DPL    (c) ADDL    (d) DLSI

**Fig.11:** Visualization of the heat map over the learned dictionary *D* by each structured DL algorithm on the UMIST face database.

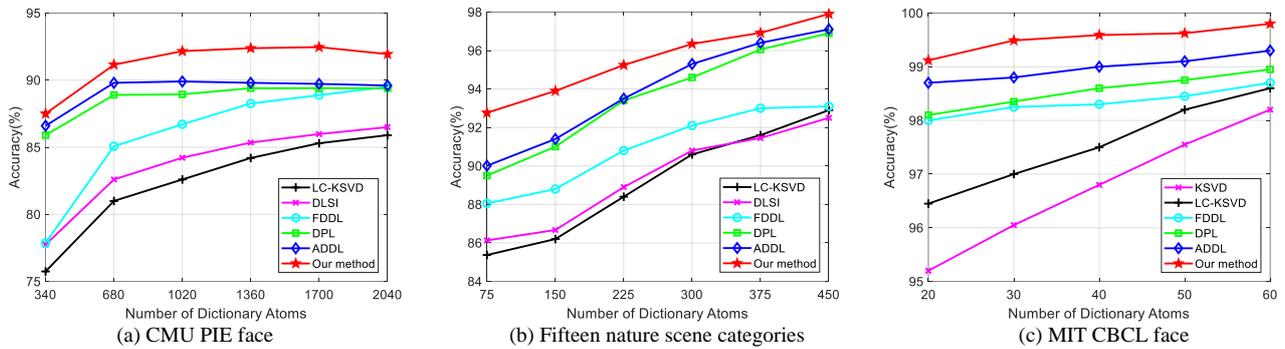

(a) CMU PIE face    (b) Fifteen nature scene categories    (c) MIT CBCL face

**Fig.11:** Quantitative recognition evaluation result of each algorithm vs. varying dictionary sizes on three real image databases.

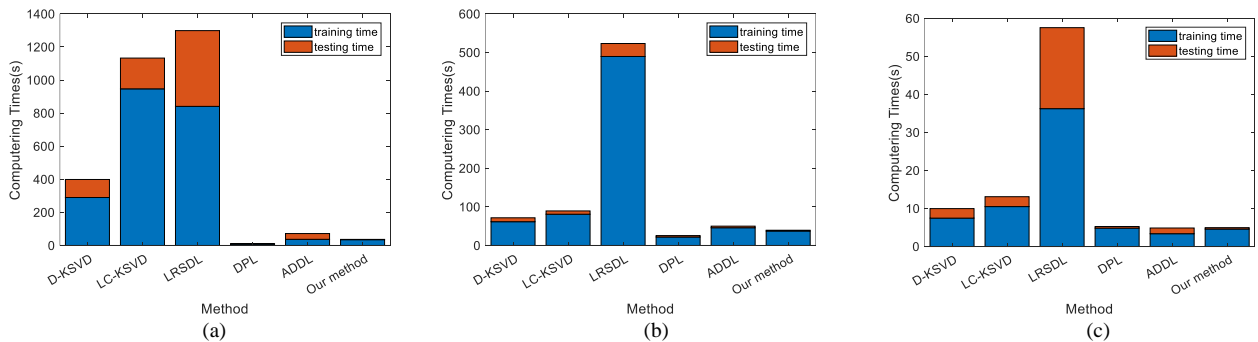

(a)    (b)    (c)

**Fig.12:** Comparison of computational time in training and test phases on (a) CMU PIE (training set: 2040 images, test set: 9514 images); (b) AR (training set:2000 images, test set: 600 images); (c) ETH80 (training set:480 images, test set: 2800 images).

training set and evaluate each method with varying sizes of dictionary, i.e., *K*=20, 30, 40, 50 and 60. The averaged results are shown in Fig.11c. We can observe that: (1) the performance of each method can be increased as the number of atoms increases in most cases; (2) RA-DPL obtains better results than its competitors. ADDL also performs well by obtaining promising results, followed by DPL and FDDL. LC-KSVD and DLSI are comparable with each other.

*G. Comparison of Computational Time*

We evaluate the training and testing time of our method and other competing methods in this study. The running time performance of our RA-DPL is mainly compared with those of DPL, ADDL, D-KSVD, LRSDL and LC-KSVD. Three databases, i.e., CMU PIE, AR and ETH80, are evaluated, and we use the same settings as *Subsection D.* We describe the averaged computational time (training and testing time) of each method over 10 runs in Fig.12, where the number of iterations is set to 20 for each method for fair comparison.

From the results, we find that: (1) DPL, ADDL and our RA-DPL are more efficient than LC-KSVD, D-KSVD and LRSDL in general. Specifically, DPL is the fastest method, followed by ADDL and our RA-DPL, respectively; (2) the testing phases of DPL and our RA-DPL are very efficient by delivering less training time than the other methods; (3) the required training time of LC-KSVD, D-KSVD and LRSDL are more than those of DPL, ADDL and our RA-DPL, since DPL, ADDL and our RA-DPL have clearly avoided using the costly $l_0/l_1$-norm for sparse representation.

## V.   CONCLUDING REMARKS

We have proposed a robust adaptive projective dictionary pair learning framework for the discriminative local sparse data representations. Our model improves the representation and discriminating abilities of existing projective dictionary pair learning from several aspects, i.e., enhancing the robust properties of the learning system to noise and corruptions in data, encouraging the coding coefficients to hold the sparse properties by efficient embedding, integrating the structured reconstruction weighting to preserve the local neighborhood within the coefficients of each class in an adaptive way, and including a discriminating function to ensure the intra-class compactness and inter-class separation over the coefficients at the same time. Due to the structured learning strategy and $l_{2,1}$-norm regularization, RA-DPL learns each sub-dictionary separately for reconstructing the data within the same class and ensures the reconstruction error to be minimized.

We have evaluated the effectiveness of our algorithm on some public databases. The investigated cases demonstrate superior performance by our RA-DPL, compared with some related models. In future, we will explore to incorporate the classifier training into the robust dictionary pair learning process. Besides, we will explore how to extend our model to the semi-supervised scenario to handle the case that the number of labeled data is limited [12][46]. Extending our method to the deep dictionary learning scenario [55-56] and evaluating it on large-scale datasets will also be discussed.

## APPENDIX I: RELATIONSHIP ANALYSIS

We illustrate some important connections to our RA-DPL.

### A. Connection to the DPL algorithm [9]

We first show that DPL is a special case of our RA-DPL. Recalling the objective function of our RA-DPL in Eq. (9), if we constrain $\beta=0, \lambda=0$, problem in Eq. (9) is reduced to

$$\langle D, P \rangle = \arg\min_{D,P} \sum_{l=1}^{c} \left\| X_l^T - X_l^T P_l^T D_l^T \right\|_{2,1} + \alpha \left( \left\| P_l \overline{X_l} \right\|_F^2 + \left\| P_l^T \right\|_{2,1} \right), \quad (25)$$
$$s.t. \ e^T D_l = e^T$$

which can be formulated as the following approximate one by expressing the $l_{2,1}$-norm with the trace equation:

$$\langle D, P, B \rangle = \arg\min_{D,P,B} \sum_{l=1}^{c} 2tr\left( \left( X_l^T - X_l^T P_l^T D_l^T \right) B \left( X_l - D_l P_l X_l \right) \right) \\ + \alpha \left( \left\| P_l \overline{X_l} \right\|_F^2 + \left\| P_l^T \right\|_{2,1} \right), \quad s.t. \ e^T D_l = e^T \quad (26)$$

where $B$ is a diagonal matrix with each diagonal entry being $B_{i,i} = 1 / \left\| 2 \left( \left( X_l^T - X_l^T P_l^T D_l^T \right)^i \right) \right\|_2$. Since $e^T D_l = e^T$ and $\|d_i\|_2^2 \leq 1$ play a very similar role, supposing that we simply use an identity matrix to replace the diagonal matrix $B$ and remove the sparse regularization on analysis dictionary, the reduced formulation is just the problem of existing DPL.

### B. Connection to the FDDL algorithm [7]

We also discussed the connection between our RA-DPL and the following simplified FDDL model [7]:

$$\langle D, S \rangle = \min_{D,S} \sum_{l=1}^{c} \left( \left\| X_l - DS_l \right\|_F^2 + \left\| X_l - D_l S_l^l \right\|_F^2 \right) + \lambda_1 \left\| S_l \right\|_1 \\ + \lambda_2 \left( \left\| S_l - M_l \right\|_F^2 - \left\| M_l - M \right\|_F^2 + \left\| S \right\|_F^2 \right) \quad , \quad (27) \\ s.t. \ \|d_i\|_2 = 1, \forall n; \ \left\| D_j S_l^j \right\|_F^2 \leq \varepsilon_f, \forall l \neq j$$

where $M_l$ and $M$ denotes the mean matrices with $m_l$ and $m$ as column vectors, $S_l^l$ is the representation coefficients of $X_l$ over $D_l$, the constraint $\left\| D_j S_l^j \right\|_F^2 \leq \varepsilon_f$ can ensure that each sub-dictionary has poor representation for other classes, $S_l^j$ is the representation coefficients of $X_l$ over $D_j$, and $\varepsilon_f$ is a small positive scalar. Recalling the problem of RA-DPL in Eq.(9), suppose that the ideal condition that $X_l^T P_l^T$ can best fit $S_l^T$ is satisfied, that is, $S_l^T = X_l^T P_l^T$, and if we further constrain parameters $\alpha=0$ and $\beta=0$, the objective function of our RA-DPL in Eq.(9) can be reduced to

$$\langle D, S \rangle = \arg\min_{D,S} \sum_{l=1}^{c} \left\| X_l^T - S_l^T D_l^T \right\|_{2,1} + \lambda \left( \left\| S_l - M_l \right\|_F^2 - N_l \left\| S_l - \overline{M_l} \right\|_F^2 \right),$$
$$s.t. \ e^T D_l = e^T, S_l \geq 0$$
$$(28)$$

which can be formulated as the following approximate one by expressing the $l_{2,1}$-norm with the trace equation:

$$\langle D, S, V \rangle = \arg\min_{D,S,V} \sum_{l=1}^{c} tr\left( \left( X_l - D_l S_l \right) V \left( X_l^T - S_l^T D_l^T \right) \right) \\ + \lambda \left( \left\| S_l - M_l \right\|_F^2 - N_l \left\| S_l - \overline{M_l} \right\|_F^2 \right), \ s.t. \ e^T D_l = e^T \quad (29)$$

where $V$ is a diagonal matrix with each diagonal entry being $V_{i,i} = 0.5 / \left\| \left( X_l^T - S_l^T D_l^T \right)^i \right\|$. By comparing Eqs.(27) and (29), we can find that RA-DPL and FDDL adopt a discriminative coefficients learning term to ensure the discriminant power of coding coefficients. The difference between FDDL and RA-DPL are twofold. First, FDDL employs a discriminative fidelity manner to gain the discriminative dictionary, while our RA-DPL uses the dictionary-pair learning mechanism to enhance the discriminative power of dictionary. Second, FDDL clearly uses the discriminative coefficients learning term $\left\| S_l - M_l \right\|_F^2 - \left\| M_l - M \right\|_F^2 + \left\| S \right\|_F^2$ to achieve the inter-class discrimination and intra-class compactness at the same time, but such operation cannot ensure the mean $m_l$ of the class $i$ to be far away from the mean $m_j (i \neq j)$ of the class $j$. In contrast, RA-DPL can potentially ensure the mean matrix $M_i$ of the class $i$ to be far away from the mean of the other classes by directly maximizing $N_l \left\| P_l X_l - \overline{M_l} \right\|_F^2$.

ACKNOWLEDGMENTS

We want to express our sincere gratitude to the anonymous referee and their comments that make our paper a higher standard. This work is partially supported by the National Natural Science Foundation of China (61672365, 61732008, 61725203, 61622305, 61871444, 61806035), High-Level Talent of the "Six Talent Peak" Project of Jiangsu Province (XYDXX-055) and the Fundamental Research Funds for the Central Universities of China (JZ2019HGPA0102). Dr. Zhao Zhang is the corresponding author of this paper.

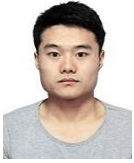

**Yulin Sun** is working toward the research degree at School of Computer Science and Technology, Soochow University, China, supervised by Dr. Zhao Zhang. His current research interests mainly include data mining, machine learning and pattern recognition. Specifically, he is very interested in desinigig advanced and robust dictionary learning algorithms for image representation and recognition.

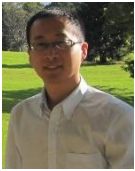

**Zhao Zhang** (SM'17- ) received the Ph.D. degree from the Department of Electronic Engineering (EE), City University of Hong Kong, in 2013. He is now a Full Professor at the School of Computer Science & School of Artificial Intelligence, Hefei University of Technology, Hefei, China. Dr. Zhang was a Visiting Research Engineer at the National University of Singapore from Feb to May 2012. He then visited the National Laboratory of Pattern Recognition (NLPR) at Chinese Academy of Sciences from Sep to Dec 2012. During Oct 2013 and Oct 2018, he was an Associate Professor at the School of Computer Science and Technology, Soochow University, Suzhou, China. His current research interests include Multimedia Data Mining & Machine Learning, Image Processing & Computer Vision. He has authored/co-authored over 80 technical papers published at prestigious journals and conferences, such as IEEE TIP (4), IEEE TKDE (6), IEEE TNNLS (4), IEEE TSP, IEEE TCSVT, IEEE TCYB, IEEE TBD, IEEE TII (2), ACM TIST, Pattern Recognition (6), Neural Networks (8), Computer Vision and Image Understanding, Neurocomputing (3), IJCAI, ACM Multimedia, ICDM (4), ICASSP and ICMR, etc. Specifically, he has published 18 regular papers in IEEE/ACM Transactions journals as the first-author or corresponding author. Dr. Zhang is serving/served as an Associate Editor (AE) for IEEE Access, Neurocomputing and IET Image Processing. Besides, he has been acting as a Senior PC member or Area Chair of ECAI、BMVC、PAKDD and ICTAI, and a PC member for 10+ popular prestigious conferences (e.g., CVPR、ICCV、IJCAI、AAAI、ACM MM、ICDM、CIKM and SDM). He is now a Senior Member of the IEEE.

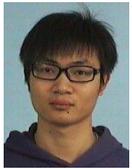

**Weiming Jiang** is working toward the research degree at School of Computer Science and Technology, Soochow University, China. His research interests include pattern recognition, machine learning and data mining. He has authored or co-authored papers published in IEEE Trans. on Neural Networks and Learning Systems (TNNLS), IEEE Transactions on Circuits and Systems for Video Technology (TCSVT), ACM International Conf. on Multimedia Retrieval (ACM ICMR), IEEE Trans. on Industrial Informatics (TII), and the IEEE International Conf. on Data Mining (ICDM).

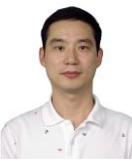

**Zheng Zhang** received the M.S and Ph.D. degrees from Harbin Institute of Technology in 2014 and 2018, respectively. His Ph.D. Thesis won the Distinguished Ph.D. Dissertation Award of The Chinese Institute of Electronics. He visited the National Laboratory of Pattern Recognition (NLPR) at the Chinese Academy of Sciences (CAS), Beijing, China. He was a Research Associate at The Hong Kong Polytechnic University, Hongkong, China, and was a Postdoctoral Research Fellow at The University of Queensland, Australia. Currently, he is an Assistant Professor at Harbin Institute of Technology, Shenzhen, China. He has authored or co-authored over 40 technical papers published at prestigious international journals and conferences, including the IEEE TPAMI, IEEE TNNLS, IEEE TIP, IEEE TCSVT, CVPR, ECCV, AAAI, IJCAI, SIGIR, and ACM Multimedia. He received the Best Paper Award from the 2014 International Conference on Smart Computing. His current research interests include machine learning and computer vision.

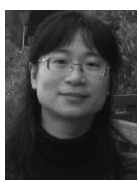

**Li Zhang** (M'08-) is currently a Professor at the School of Computer Science and Technology, Soochow University, Suzhou, China. She obtained her Bachelor and PhD degrees from Xidian University in 1997 and 2002, respectively. She was a postdoc researcher at the Shanghai Jiaotong University from 2003 to 2005. His research interests include pattern recognition, machine learning, and data mining. She has authored/co-authored more than 90 technical papers published at prestigious international journals and conferences, including IEEE TNNLS, IEEE TSP, IEEE Trans. SMC Part B, Information Sciences, Pattern Recognition, Neural Networks, etc.

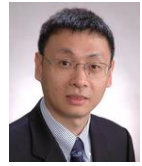

**Shuicheng Yan** (F'16- ) received the Ph.D. degree from the School of Mathematical Sciences, Peking University, in 2004. He is currently the Dean's Chair Associate Professor at National University of Singapore, and also the chief scientist of Qihoo/360 company. Dr. Yan's research areas include machine learning, computer vision and multimedia, and he has authored/ co-authored hundreds of technical papers over a wide range of research topics, with Google Scholar citation over 20,000 times and H-index 66. He is ISI Highly-cited Researcher of 2014-2016. He is an associate editor of IEEE Trans. Knowledge and Data Engineering, IEEE Trans. on Circuits and Systems for Video Technology (IEEE TCSVT) and ACM Trans. Intelligent Systems and Technology (ACM TIST). He received the Best Paper Awards from ACM MM'12 (demo), ACM MM'10, ICME'10 and ICIMCS'09, the winner prizes of classification task in PASCAL VOC 2010-2012, the winner prize of the segmentation task in PASCAL VOC 2012, 2010 TCSVT Best Associate Editor (BAE) Award, 2010 Young Faculty Research Award, 2011 Singapore Young Scientist Award and 2012 NUS Young Researcher Award. He is a Fellow of the IEEE and IAPR.

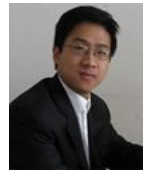

**Meng Wang** is a Professor in the Hefei University of Technology, China. He received the B.E. degree and Ph.D. degree in the Special Class for the Gifted Young and signal and information processing from the University of Science and Technology of China (USTC), Hefei, China, respectively. His current research interests include multimedia content analysis, search, mining, recommendation, and large-scale computing. He has authored 6 book chapters and over 100 journal and conference papers in these areas, including IEEE TMM, TNNLS, TCSVT, TIP, TOMCCAP, ACM MM, WWW, SIGIR, ICDM, etc. He received the paper awards from ACM MM 2009 (Best Paper Award), ACM MM 2010 (Best Paper Award), MMM 2010 (Best Paper Award), ICIMCS 2012 (Best Paper Award), ACM MM 2012 (Best Demo Award), ICDM 2014 (Best Student Paper Award), PCM 2015 (Best Paper Award), SIGIR 2015 (Best Paper Honorable Mention), IEEE TMM 2015 (Best Paper Honorable Mention), and IEEE TMM 2016 (Best Paper Honorable Mention). He is the recipient of ACM SIGMM Rising Star Award 2014. He is/has been an Associate Editor of IEEE Transactions on Knowledge and Data Engineering (TKDE), IEEE Transactions on Neural Networks and Learning Systems (TNNLS) and IEEE Transactions on Circuits and Systems for Video Technology (TCSVT). He is a senior member of the IEEE.